\documentclass{article}

\usepackage{microtype}
\usepackage{graphicx}
\usepackage{booktabs} 
\usepackage{amsmath,amssymb,amsfonts}
\usepackage{algorithmic}
\usepackage{textcomp}
\usepackage[table,xcdraw]{xcolor}
\usepackage{enumitem}
\usepackage{multirow}   
\usepackage{makecell}   
\usepackage[caption=false,font=footnotesize]{subfig}
\definecolor{mygreen}{RGB}{0,128,0}
\usepackage[font=small,labelfont=bf,skip=3pt]{caption}

\usepackage[hyperfootnotes=false]{hyperref}

\usepackage[accepted]{mlsys2026}
\usepackage{fvextra}
\DefineVerbatimEnvironment{breakableverbatim}{Verbatim}{breaklines=true,breakanywhere=true}

\AddToShipoutPictureBG*{
  \AtPageUpperLeft{
    \hspace*{\dimexpr\paperwidth-16pt\relax}
    \raisebox{-68pt}{
      \llap{
        \href{https://www.acm.org/publications/policies/artifact-review-and-badging-current}{
        \includegraphics[height=65pt]{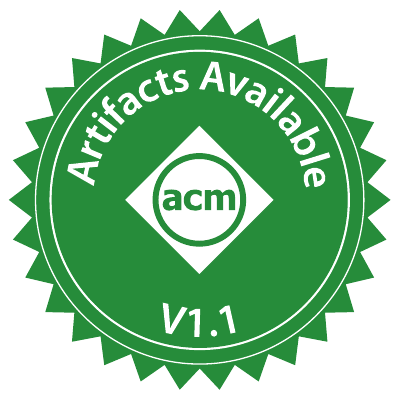}}
        \hspace{1pt}
        \href{https://www.acm.org/publications/policies/artifact-review-and-badging-current}{
        \includegraphics[height=65pt]{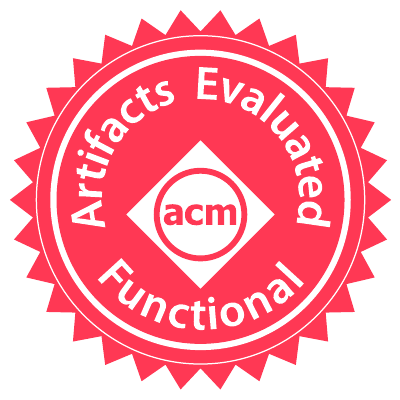}}
      }
    }
  }
}

\mlsystitlerunning{IntAttention: A Fully Integer Attention Pipeline for Efficient Edge Inference}

\begin{document}


\twocolumn[
  \mlsystitle{IntAttention: A Fully Integer Attention Pipeline for Efficient Edge Inference}



  \mlsyssetsymbol{equal}{*}

  \begin{mlsysauthorlist}
    \mlsysauthor{Wanli Zhong}{sustech}
    \mlsysauthor{Haibo Feng}{sustech,pcl}
    \mlsysauthor{Zirui Zhou}{sustech}
    \mlsysauthor{Hanyang Peng}{pcl}
    \mlsysauthor{Shiqi Yu}{sustech}
  \end{mlsysauthorlist}

  \mlsysaffiliation{sustech}{Department of Computer Science and Engineering, Southern University of Science and Technology, Shenzhen, China}
  \mlsysaffiliation{pcl}{Peng Cheng Laboratory, Shenzhen, China}

  \mlsyscorrespondingauthor{Shiqi Yu}{yusq@sustech.edu.cn}

  \mlsyskeywords{Machine Learning, MLSys}

  \vskip 0.3in

  \begin{abstract}
    Deploying Transformer models on edge devices is limited by latency and energy budgets. While INT8 quantization effectively accelerates the primary matrix multiplications, it exposes the softmax-related path as the dominant bottleneck. This stage incurs a costly $\mathrm{dequantize}\rightarrow\mathrm{softmax}\rightarrow\mathrm{requantize}$ detour, which can account for up to 65\% of total attention latency and disrupts the end-to-end integer dataflow critical for edge hardware efficiency. To address this limitation, we present \emph{IntAttention}, the first fully integer attention pipeline that serves as a training-free drop-in replacement. At the core of our approach lies \emph{IndexSoftmax}, a hardware-friendly operator that replaces floating-point exponentials entirely within the integer domain. \emph{IntAttention} integrates sparsity-aware clipping, a 32-entry lookup-table approximation, and direct integer normalization, thereby eliminating datatype conversion overhead along the attention path. Experiments on Armv8 CPUs show that our method achieves up to \textbf{3.7$\times$} speedup and \textbf{61\%} energy reduction over FP16 baselines, and up to \textbf{2.0$\times$} speedup over conventional INT8 attention pipelines. Across diverse language and vision models, as well as additional reasoning and long-context evaluations, \emph{IntAttention} maintains strong overall fidelity and demonstrates a more favorable trade-off than existing LUT-based softmax approximations. The code is available at \url{https://github.com/WanliZhong/IntAttention}
  \end{abstract}
]






\printAffiliationsAndNotice{}  

\section{Introduction}
\label{sec:introduction}
Transformer-based models achieve state-of-the-art performance across natural language processing \cite{vaswani2017attention}, computer vision \cite{dosovitskiy2020image}, and multimodal tasks \cite{radford2021learning}. The core mechanism, multi-head self-attention, exhibits quadratic time and memory complexity with respect to sequence length ($O(L^2)$). As the context length increases, attention becomes the dominant inference cost, substantially increasing latency and memory usage. In autoregressive language models, the prefill phase largely determines the time-to-first-token (TTFT), as the full key–value cache must be computed before generation begins \cite{kwon2023efficient}. Consequently, long prompts are computationally expensive, even though subsequent decoding is relatively fast.

\begin{figure}[t]
  \centering
  \hspace{0.02\textwidth}
  \subfloat[Quantized Attention]{
    \includegraphics[width=0.18\textwidth]{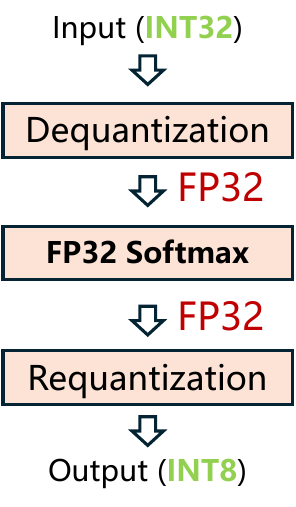}
  }\hfill
  \subfloat[IntAttention (Ours)]{
    \includegraphics[width=0.18\textwidth]{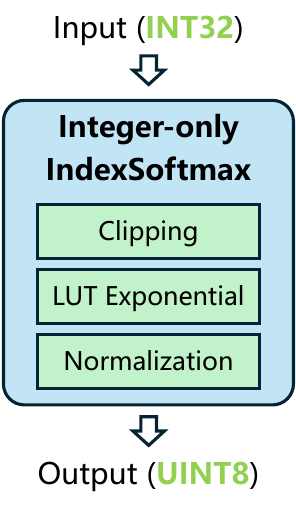}
  }
  \hspace*{0.02\textwidth}
  \caption{Conventional quantized attention falls back to floating point in the softmax stage, whereas proposed \textbf{IntAttention} maintains an end-to-end integer dataflow from $\mathbf{QK}^\top$ to $\mathbf{PV}$.}
  \label{fig:diff_qa_ia}
\end{figure}

Recent advances in compact and specialized models have accelerated the transition toward on-device inference. For instance, Google’s Gemma3-270M \cite{team2025gemma} model targets energy-efficient deployment on smartphones, while studies such as QuestA demonstrate that reinforcement learning-based question augmentation can elevate a 1.5B model to the performance level of 32B models on multiple reasoning benchmarks \cite{li2025questa}. This migration shifts inference from cloud servers to mobile and embedded processors, where end-to-end latency and energy efficiency become primary constraints. Consequently, optimizing attention, particularly by reducing the computational cost of long-context processing, has become crucial for deploying practical large language models (LLMs) on edge hardware.

Previous studies commonly use low-precision floating-point formats such as BF16, FP16, FP8, and even FP4 \cite{micikevicius2022fp8, zhang2025sageattention3, shah2024flashattention}. However, these formats are neither universally supported nor energy-efficient on commodity edge hardware, which typically offers highly optimized integer computation pathways. Consequently, integer quantization, especially INT8, is the most practical and effective optimization strategy \cite{jacob2018quantization, dettmers2022gpt3}. Motivated by this hardware constraint, we propose \textbf{IntAttention}. To the best of our knowledge, this work presents the \textbf{first fully integer attention pipeline that serves as a training-free drop-in replacement.} Designed for off-the-shelf edge processors, it executes attention entirely in the integer domain, eliminating redundant dequantization and requantization steps. Consequently, it functions as a drop-in replacement for conventional quantized attention within transformer-based inference pipelines.

Achieving an end-to-end integer attention pipeline is non-trivial. After applying dynamic INT8 quantization to the $\mathbf{QK}^\top$ and $\mathbf{PV}$ matrix multiplications, the remaining path that forms and applies attention weights becomes the dominant bottleneck. The standard softmax requires floating-point exponentials, row-wise normalization, and repeated data-format conversions.
On edge processors, this $\mathrm{dequantize}\rightarrow\mathrm{softmax}\rightarrow\mathrm{requantize}$ detour can account for up to \textbf{65\%} of the attention latency once the surrounding GEMMs are quantized as shown in \autoref{fig:softmax_ratio}, thereby eroding much of the benefit of integer GEMMs acceleration.

To mitigate this bottleneck, we introduce \emph{IndexSoftmax}, a lightweight approximation that replaces costly exponential evaluations with a compact lookup table and performs max-subtraction and clipping in the integer domain. \emph{IndexSoftmax} preserves the relative structure of the attention scores and eliminates most per-element floating-point work.

Building on this, we integrate \emph{IndexSoftmax} with integer normalization and direct requantization of the probability tensor.
The resulting pipeline, \textbf{IntAttention}, takes INT32 logits from the $\mathbf{QK}^\top$ GEMM, produces a quantized attention tensor $\hat{\mathbf{P}}$ in UINT8, and feeds $\hat{\mathbf{P}}$ into the integer value-aggregation kernel.
This design eliminates the $\mathrm{dequantize}\rightarrow\mathrm{softmax}\rightarrow\mathrm{requantize}$  detour and preserves an end-to-end integer dataflow.
\autoref{fig:diff_qa_ia} contrasts a conventional quantized pipeline, which falls back to floating point in the softmax stage, with \textbf{IntAttention}, which remains integer from input to output.

To summarize, this work makes the following contributions:
\begin{itemize}
  \item We introduce \emph{IndexSoftmax}, a lookup-table-based approximation to softmax that is compatible with integer execution, and integrate it with integer normalization and probability quantization to form \textbf{IntAttention}.
  \item We show that \textbf{IntAttention} runs on off-the-shelf edge processors, delivering up to \textbf{3.7$\times$} speedup and up to \textbf{61\%} lower energy consumption than an FP16 baseline, while maintaining strong overall fidelity on both language and vision models.
\end{itemize}

\section{Background and Motivation}
\subsection{Attention and Dynamic Quantization}
\label{sec:attn-and-dq}

\paragraph{Scaled Dot-Product Attention}
We first recall the standard attention mechanism, which underlies our design.
Let $\mathbf{Q}, \mathbf{K}, \mathbf{V} \in \mathbb{R}^{L \times d}$ denote the query, key, and value with sequence length $L$ and feature dimension per token $d$ \cite{vaswani2017attention}.
The standard attention computes
\begin{equation}
  \mathbf{A} = \mathbf{Q}\mathbf{K}^\top,\quad
  \mathbf{P} = \operatorname{softmax}\!\left(\frac{\mathbf{A}}{\sqrt{d}}\right),\quad
  \mathbf{O} = \mathbf{P}\mathbf{V}.
  \label{eq:attention_re}
\end{equation}

\paragraph{Dynamic Quantizing \texorpdfstring{$\mathbf{Q},\mathbf{K},\mathbf{V}$}{Q,K,V}}
To reduce latency and memory traffic on edge hardware, we adopt per-tensor symmetric INT8 quantization with zero point fixed at $0$ for the three inputs $\mathbf{Q}, \mathbf{K}, \mathbf{V}$ \cite{jacob2018quantization}.
For a tensor $\mathbf{X}$, let $\hat{\mathbf{X}}=\operatorname{quant}(\mathbf{X})$; the scale factor and the quantized tensor are
\begin{equation}
  s_X = \frac{\max(|\mathbf{X}|)}{127},
\end{equation}
\begin{equation}
  \operatorname{quant}(\mathbf{X}) = \operatorname{clamp}\!\Big(\Big\lfloor \frac{\mathbf{X}}{s_X} \Big\rceil, -127, 127 \Big),
  \label{eq:quant}
\end{equation}
which enables low-precision matrix multiplications while preserving a simple dequantization model $\mathbf{X} \approx s_X \hat{\mathbf{X}}$, where $\hat{\mathbf{X}} \in \{-127,\ldots,127\}^{L\times d}$.
Applying \autoref{eq:quant} to $\mathbf{Q}$, $\mathbf{K}$, and $\mathbf{V}$ yields their quantized counterparts $\hat{\mathbf{Q}}, \hat{\mathbf{K}}, \hat{\mathbf{V}}$ and the corresponding scales $s_Q, s_K, s_V$.

\paragraph{Integer accumulation and scaling.}
After quantization, the attention logits are computed fully in the integer domain using INT8$\times$INT8 multiplications with INT32 accumulation:
\begin{equation}
  \hat{\mathbf{A}} = \hat{\mathbf{Q}}\hat{\mathbf{K}}^\top,\quad
  \alpha = \frac{s_Q s_K}{\sqrt{d}},\quad
  \mathbf{A} \approx \alpha \, \hat{\mathbf{A}} .
  \label{eq:logit-scale}
\end{equation}
Here $\alpha$ rescales integer logits to the floating-point range.
Substituting $\mathbf{V} \approx s_V \hat{\mathbf{V}}$ into $\mathbf{O}=\mathbf{P}\mathbf{V}$ gives
\begin{equation}
  \mathbf{O} = \mathbf{P}\mathbf{V} \approx s_V \, \mathbf{P}\hat{\mathbf{V}} .
  \label{eq:pv-link}
\end{equation}

With these definitions, the two heavy matrix multiplications ($\mathbf{QK}^\top$ and $\mathbf{PV}$) are executed in integer arithmetic.
SageAttention demonstrates that carefully engineered INT8 kernels yield significant throughput gains with minimal loss of accuracy across diverse models \cite{zhang2024sageattention}.
SageAttention2 further demonstrates that pushing similarity computation to INT4 for $\mathbf{Q}$ and $\mathbf{K}$ while keeping a slightly higher precision on the value path can remain near lossless and provides additional speedups on modern GPUs \cite{zhang2024sageattention2}.

\subsection{Emerging Bottleneck in Quantized Attention Pipelines}
\label{sec:softmax-bottleneck}
\paragraph{Numerically stable softmax.}
We adopt the standard maximum subtraction strategy to ensure numerical stability in the exponential.
Given $\mathbf{A}\in\mathbb{R}^{L\times L}$, define the row-wise maximum vector
$\mathbf{m}=\operatorname{rowMax}(\mathbf{A})$.
The stable softmax can then be written compactly as
\begin{equation}
  \mathbf{P}
  = \frac{\exp\big(\mathbf{A}-\mathbf{m}\big)}
  {\operatorname{rowSum}\left(\exp\big(\mathbf{A}-\mathbf{m}\big)\right)}
  \label{eq:safe-softmax}
\end{equation}
where both \emph{rowMax} and \emph{rowSum} operate along the row dimension.
This transformation preserves the exact softmax while constraining all exponential inputs to $(-\infty,0]$,
thus preventing numerical overflow and improving stability.

\paragraph{Cost drivers on edge hardware.}
The $O(L^2)$ complexity of softmax persists even when $\mathbf{QK}^\top$ and $\mathbf{PV}$ are accelerated.
On edge hardware with limited thread-level parallelism, the exponential and the division dominate latency.
A single $\exp(\cdot)$ typically expands to tens of floating point operations per element, and the normalization still incurs row-wise divisions.
In quantized pipelines, this cost is further amplified because INT32 logits must be dequantized to FP32 before softmax, and the resulting probabilities must be requantized for the value projection, which interrupts an otherwise contiguous integer dataflow.

\begin{figure}[!t]
  \centering
  \includegraphics[width=0.48\textwidth]{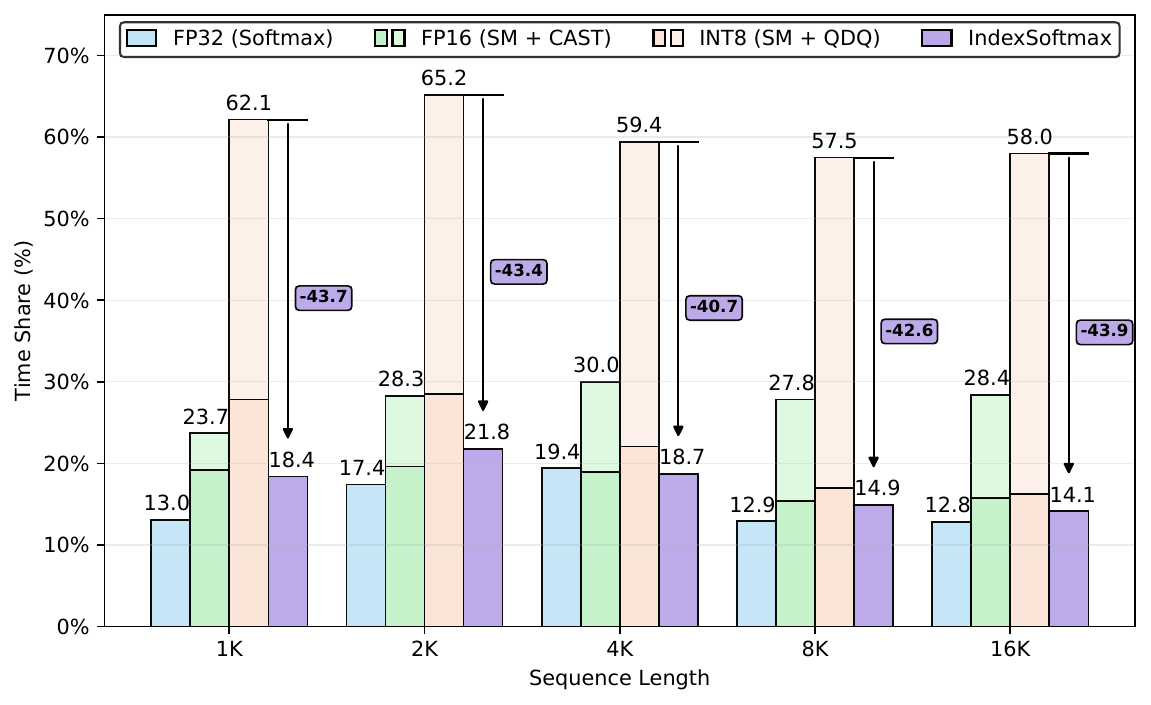}
  \vspace{-8mm}
  \caption{Breakdown of time share for the $\mathrm{dequantize}\rightarrow\mathrm{softmax}\rightarrow\mathrm{requantize}$ path across different precisions. Once GEMMs are accelerated to INT8, this path emerges as the dominant latency and becomes the next optimization target.}
  \label{fig:softmax_ratio}
  \vspace{-3mm}

\end{figure}

\paragraph{Measured breakdown and implication.}
\autoref{fig:softmax_ratio} reports the measured time share of the $\mathrm{dequantize}\rightarrow\mathrm{softmax}\rightarrow\mathrm{requantize}$ path.
In FP32, the share is about 13\% to 19\% across sequence lengths.
With FP16 GEMMs it increases to about 23\% to 30\% yet remains secondary.
After switching GEMMs to INT8, their latency drops while the softmax path is essentially unchanged, so its share rises to \textbf{57\% to 65\%} and becomes the dominant cost.
Therefore, once the multiplications are quantized, the probability normalization path is the next component that must be optimized to unlock further end to end speedups.

\paragraph{Context in prior work.}
Multiple GPU–oriented efforts have shown that softmax becomes the limiting stage once the surrounding matrix multiplications are heavily optimized. FlashAttention related kernels tile queries and keys, fuse attention with online softmax, and aggressively reduce memory traffic \cite{dao2022flashattention, dao2023flashattention}. FlashAttention-3 goes further by driving GEMMs with FP8 Tensor Cores and then hiding the dominant softmax cost using warp specialization and ping–pong scheduling that overlaps GEMM and softmax through double buffering \cite{shah2024flashattention}. TurboAttention extends this line of work by addressing not only the softmax bottleneck but also the dequantization overhead that arises in quantized attention. It unifies these optimizations through \emph{FlashQ}, which enables quantized execution of matrix multiplications, and a sparsity-based softmax approximation that avoids FP32 dequantization during exponentiation\cite{kang2024turboattention}.

These results confirm that the softmax and quantization-related path is a real bottleneck. However, all of these optimizations rely on massive GPU parallelism and specialized floating-point hardware. On edge devices, which lack high-throughput floating-point units and deep warp-level concurrency but provide efficient integer units, the same path remains largely scalar and costly. This motivates the need for a lightweight, training-free, and integer-friendly softmax replacement to further accelerate attention inference on edge hardware.

\begin{figure*}
  \centering
  \includegraphics[width=0.95\textwidth]{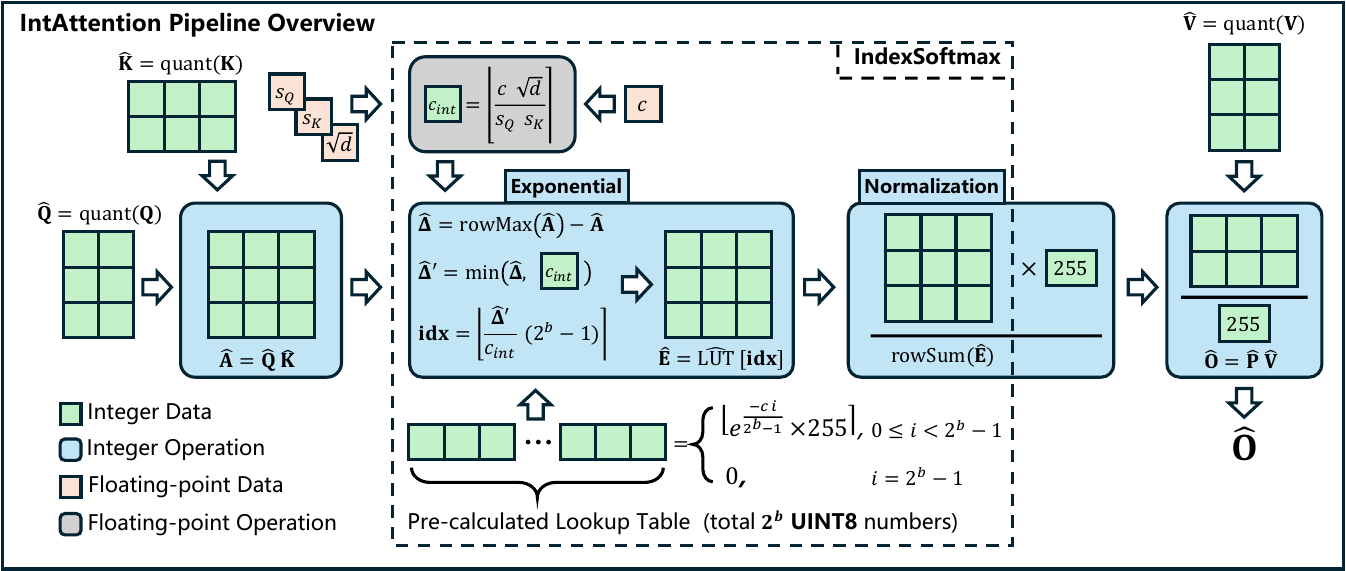}
  \vspace{-3mm}
  \caption{Overview of the proposed \textbf{IntAttention} pipeline.}
  \label{fig:pipeline}
  \vspace{-3mm}
\end{figure*}

\subsection{Acceleration Strategies}
\label{sec:prior-work-critique}
As the softmax and its associated floating-point conversion overhead have become the dominant bottlenecks in quantized attention, recent research has focused on accelerating this stage through three main strategies:

\paragraph{Hardware-oriented softmax co-design.}
Approaches such as Softermax and ConSmax redesign the softmax operator alongside dedicated accelerator logic. Softermax replaces \(e^x\) with \(2^x\) and uses fixed-point integer shifters for exponentiation and normalization \cite{stevens2021softermax}. ConSmax removes explicit max-finding and normalization by training fixed scaling constants, allowing inference to be implemented with table lookups and multiplications \cite{liu2024consmax}. These co-designs achieve high throughput and energy savings, but only work on specialized hardware and require operator changes, which limits their adoption on generic edge processors.

\paragraph{Input-aware quantization and LUT-based softmax.}
Methods like EXAQ and TurboAttention accelerate softmax without changing model operators or hardware. EXAQ determines dynamic optimal clipping ranges to quantize attention scores to as low as 3 bits \cite{shkolnik2024exaq}. TurboAttention uses a small LUT for the integer part of the exponent and a 3rd-order polynomial for the fractional part, plus sparsification of negligible exponentials \cite{kang2024turboattention}. These techniques eliminate heavy floating-point exponent operations and can be applied directly in attention inference, but the normalization step (sum and divide) typically remains in high-precision arithmetic, so the dataflow remains mixed-precision and still burdens edge CPUs.

\paragraph{Integer-only softmax in Transformer quantization.}
Fully integer softmax schemes are integrated into integer-quantized Transformer pipelines. I-BERT uses low-order integer polynomials and iterative integer refinements to approximate softmax \cite{kim2021bert}. I-ViT introduces \emph{Shiftmax}, expressing the exponential via bit shifts and additions \cite{li2023vit}. I-LLM proposes \emph{DI-ClippedSoftmax}, performing clipping and scaling entirely in integer form \cite{hu2024llm}. Although these methods deliver a true integer dataflow, they usually depend on quantization-aware training or calibration to recover accuracy, and add runtime overhead for scale/clip estimation factors that limit their seamless deployment on edge devices.

In summary, hardware co-design approaches deliver very high efficiency but depend on specialized logic and retraining, which limits portability. Input-aware quantization and LUT-based methods are more deployable, but they typically leave the normalization step in floating-point or high-precision calculation, so the dataflow is still mixed-precision. Integer-only softmax methods promise a fully integer path, but they require model adaptation through fine-tuning or reconstruction. None of these families removes the full $\mathrm{dequantize}\rightarrow\mathrm{softmax}\rightarrow\mathrm{requantize}$ loop in a way that is simultaneously fully integer, training-free, and directly deployable on commodity edge CPUs.

\subsection{Motivation and Design Goals}
\label{sec:motivation}

The breakdown in \autoref{sec:softmax-bottleneck} shows that once the matrix multiplications in attention are quantized and accelerated, the dominant latency on edge processors comes from the remaining $\mathrm{dequantize}\rightarrow\mathrm{softmax}\rightarrow\mathrm{requantize}$ path. Prior work alleviates parts of this path, but always at the cost of at least one key property: it either assumes custom hardware, falls back to floating point in the normalization step, or requires model retraining. As a result, the practical end-to-end gain for real deployment remains limited.

Our objective is to remove this bottleneck in a way that is directly usable in existing quantized attention pipelines. This leads to four design goals:

\begin{enumerate}
  \item \textbf{Integer execution.} All stages of attention, including the analogue of exponentiation and the row-wise normalization, must run in integer arithmetic. This allows the computation to fully exploit the efficient integer units that are already available on edge hardware, instead of invoking slower floating-point paths.

  \item \textbf{Drop-in deployment.} The method must serve as a drop-in replacement for standard attention in pre-trained models. It should not require quantization-aware training, fine-tuning, or structural changes. This makes it directly adoptable in large existing models and gives it immediate deployment value.

  \item \textbf{Portable efficiency.} The implementation must rely only on common integer primitives such as add, multiply, shift, and indexed lookup, and must parallelize cleanly on SIMD-style cores (for example ARM NEON). It should not introduce extra global passes for per-input statistics. This keeps the method practical on a wide range of commodity devices.

  \item \textbf{Fidelity under acceleration.} The operator must deliver a clear latency and energy advantage over floating-point softmax, while maintaining accuracy close to the original FP16 attention. In other words, efficiency gains cannot come at the expense of unacceptable degradation.
\end{enumerate}

\section{IntAttention}
\label{sec:intattention}
\textbf{IntAttention} transforms the conventional quantized attention block into a true integer-domain pipeline, removing the $\mathrm{dequantize}\rightarrow\mathrm{softmax}\rightarrow\mathrm{requantize}$ detour that dominates latency on edge processors. By preserving a contiguous integer path from the $\mathbf{QK}^\top$ logits to the $\mathbf{PV}$ multiplication, \textbf{IntAttention} executes entirely on commodity integer units, requires no model retraining, and can be used as a drop-in replacement in existing quantized attention inference.

At the kernel of \textbf{IntAttention} is \emph{IndexSoftmax}, whose core operation is integer clipping followed by lookup-table based exponent approximation. The resulting $\mathrm{UINT8}$ attention map \(\hat{\mathbf{P}}\) feeds directly into the integer $\mathbf{PV}$ kernel, so no floating-point computation appears on the runtime path.

Implementation of \emph{IndexSoftmax} relies on three tightly coupled mechanisms: integer-domain clipping, LUT exponentials, and integer scale normalization, which are designed and tuned together rather than as independent modules. This coupling minimizes extra passes or global statistics, preserves parallelism on SIMD-style integer units, and yields substantial reductions in latency and energy while maintaining strong overall fidelity.

An overview of the proposed \emph{IntAttention} pipeline is illustrated in \autoref{fig:pipeline}. The following subsections detail each mechanism and the integration choices that enable efficient, portable integer attention.

\subsection{IndexSoftmax}

\begin{figure}
  \centering
  \includegraphics[width=0.48\textwidth]{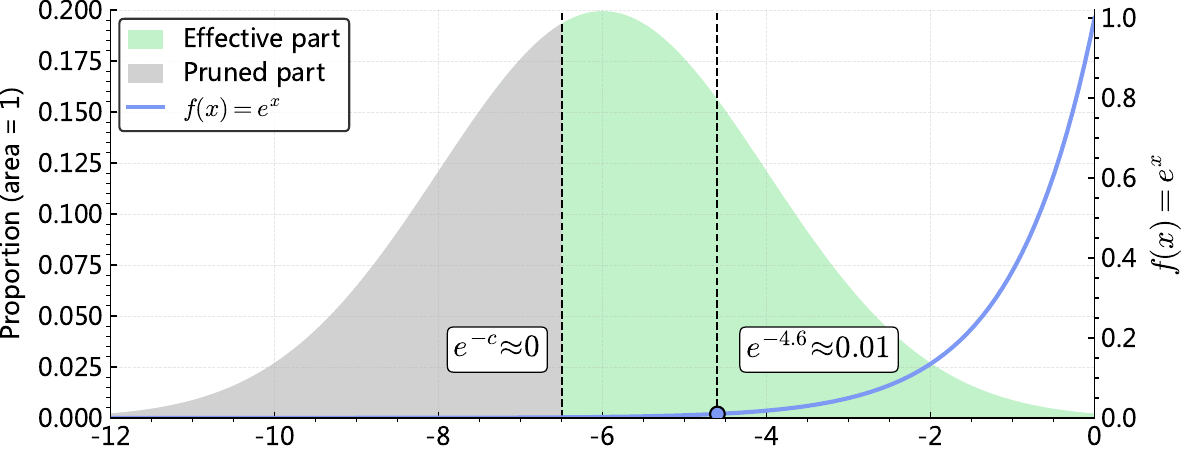}
  \vspace{-7mm}
  \caption{The exponential activation in softmax is dominated by a small subset of high logits, while most logits lie in the near-zero region and contribute negligibly to normalization.}
  \label{fig:exp_distribution}
\end{figure}

\paragraph{Integer-Domain Clipping via Sparsity-Aware Pruning}
\label{sec:clipping}
The exponential in Softmax exhibits an inherent \emph{sparsity}: as inputs decrease, $\exp(\cdot)$ rapidly approaches zero. In practice shown in \autoref{fig:exp_distribution}, a small subset of high valued logits dominates the normalization term, while the majority contribute negligibly. Evaluating exponentials for these near-zero terms wastes arithmetic and increases memory traffic, especially on edge devices. To exploit this, we introduce an \textbf{integer-domain clipping} mechanism that removes low-importance logits before the exponential approximation. Unlike floating-point sparse/efficient attention variants, our method stays entirely in the integer domain and avoids type conversions.

Formally, given integer logits $\hat{\mathbf{A}}\!\in\!\mathbb{Z}^{L\times L}$, we apply row-wise max-subtraction for stability:
\begin{equation}
  \hat{\boldsymbol{\Delta}} \;=\; \operatorname{rowMax}(\hat{\mathbf{A}}) \;-\; \hat{\mathbf{A}},
\end{equation}
which yields nonnegative distances from the dominant value in each row. We then use a quantization aligned clipping threshold $c_{\text{int}}$, derived from the clipping threshold $c$ via the quantization scale factor in \autoref{eq:logit-scale}:
\begin{equation}
  \label{eq:clipping_thr}
  c_{\text{int}} = \operatorname{round}\Big(\frac{c}{\alpha}\Big)
  = \operatorname{round}\Big(\frac{c\,\sqrt{d}}{s_Q\,s_K}\Big).
\end{equation}
Clipping is performed elementwise:
\begin{equation}
  \label{eq:int_clipping}
  \hat{\boldsymbol{\Delta}}' = \min\big(\hat{\boldsymbol{\Delta}},\, c_{\text{int}}\big),
\end{equation}
so entries whose contributions to $\exp(-\alpha\,\hat{\boldsymbol{\Delta}})$ are negligible are saturated at $c_{\text{int}}$.

To match LUT-based exponentiation, we adopt the sign convention $\mathbf{m}-\mathbf{A}$ (rather than $\mathbf{A}-\mathbf{m}$ in \autoref{eq:safe-softmax}), ensuring all arguments to $\exp(-x)$ are nonnegative and lie within $[0,c]$. Combined with clipping, this confines the exponential evaluation to a compact, table-friendly domain.

Overall, integer-domain clipping provides two benefits: (i) it removes redundant work on near-zero contributions, reducing arithmetic and bandwidth, and (ii) it establishes a quantization consistent, bounded range for integer-domain exponentiation. These properties lay the groundwork for an efficient, fully integer Softmax pipeline that is both sparsity-aware and deployment friendly on low-power accelerators.

\paragraph{Efficient Exponential Approximation Using Lookup Tables}
\label{sec:lut-exp}
Evaluating $\exp(\cdot)$ is costly in quantized inference. Classical implementations use iterative or polynomial schemes (e.g., Padé or Taylor series) that require multiple floating-point operations. On GPUs this cost can be amortized by massive parallelism, but on edge devices, where bandwidth and instruction latency dominate, $\exp(\cdot)$ often becomes the bottleneck once the matrix multiplications are quantized. The result is a fast integer $\mathbf{QK}^\top$ followed by a floating-point Softmax that breaks the integer dataflow and limits further speedup.

We address this by replacing the exponential stage with a \emph{table-driven} surrogate. After integer-domain clipping defined in \autoref{eq:int_clipping}, all inputs to $\exp(-x)$ lie in a finite interval $[0,c]$. Over this range the function is bounded, so a fixed resolution discretization provides an effective and simple approximation. We therefore precompute a \emph{fixed} lookup table with $2^{b}$ entries,

\begin{equation}
  \mathrm{LUT}[i] =
  \begin{cases}
    \exp \left(-\dfrac{c \, i}{2^{b}-1}\right), & 0 \le i < 2^{b}-1, \\
    0, & i = 2^{b}-1.
  \end{cases}
  \label{eq:lut_re}
\end{equation}

and map clipped integer distances to indices by a linear rescaling,

\begin{equation}
  \boldsymbol{\mathrm{idx}} = \Big\lfloor\frac{\hat{\boldsymbol{\Delta}}'}{c_{\text{int}}}\left(2^{b}-1\right)\Big\rceil,
\end{equation}

The exponential surrogate is then obtained by a gather:

\begin{equation}
  \tilde{\boldsymbol{\mathrm{E}}} = \mathrm{LUT}\big[\boldsymbol{\mathrm{idx}}\big] \approx \exp\big(-\alpha\,\hat{\boldsymbol{\Delta}}'\big),
\end{equation}

Among LUT-only methods, the closest is EXAQ, which uses a \emph{dynamic} clipping rule based on per-tensor standard deviation statistics together with ultra-low LUT resolutions ($b\!\in\!\{2,3\}$) \cite{shkolnik2024exaq}. This adds global reductions and control overhead that are expensive on edge devices. In contrast, we adopt \emph{fixed} hyperparameters $(c,b)$ selected offline. Empirically shown in \autoref{fig:hyper_joint}, performance is insensitive to $c$ within a practical range, and modestly increasing $b$ to a moderate table size has a negligible runtime impact while clearly improving approximation accuracy. As long as the table remains moderate, lookup latency is effectively constant, so pursuing extremely small tables offers little real benefit but degrades fidelity. A moderate, fixed-resolution LUT achieves a stronger balance between accuracy and efficiency for integer attention on edge hardware.

\subsection{LUT Rebuild and Integer Scale Normalization}
\label{sec:int-norm}

\begin{figure}
  \centering
  \includegraphics[width=0.48\textwidth]{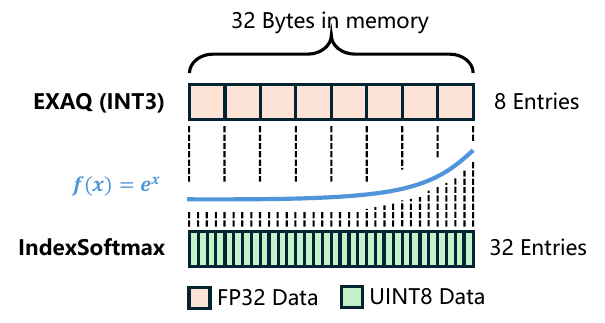}
  \vspace{-7mm}
  \caption{Under the same memory budget, \textbf{IndexSoftmax} provides 4$\times$ higher LUT resolution than EXAQ, enabling higher-fidelity exponential approximation without dynamic clipping or global statistics.}
  \label{fig:exaq_vs_idxsoftmax}
\end{figure}

Quantization of the probability matrix $\mathbf{P}$ has a crucial impact on the final attention output.
While prior methods often scale probabilities by $\times 127$ and store them in signed INT8,
we adopt an unsigned UINT8 formulation scaled by $\times 255$,
which fully utilizes the available range and improves numerical smoothness during normalization.
This design allows the softmax path to remain entirely in the integer domain, with both the lookup table and output probabilities quantized to UINT8.
Because the precision of $\mathbf{P}$ is inherently limited by its 8-bit representation,
using excessively precise floating-point lookup tables brings little benefit.
Therefore, our exponential lookup table is also quantized to UINT8,
so that each entry is compact yet expressive enough to represent the clipped exponential curve.

Over the clipped interval $[0,c]$, the floating-point table is linearly mapped to integers table:

\begin{figure*}[!t]
  \centering
  \begin{minipage}[t]{0.48\textwidth}
    \centering
    \includegraphics[width=\linewidth]{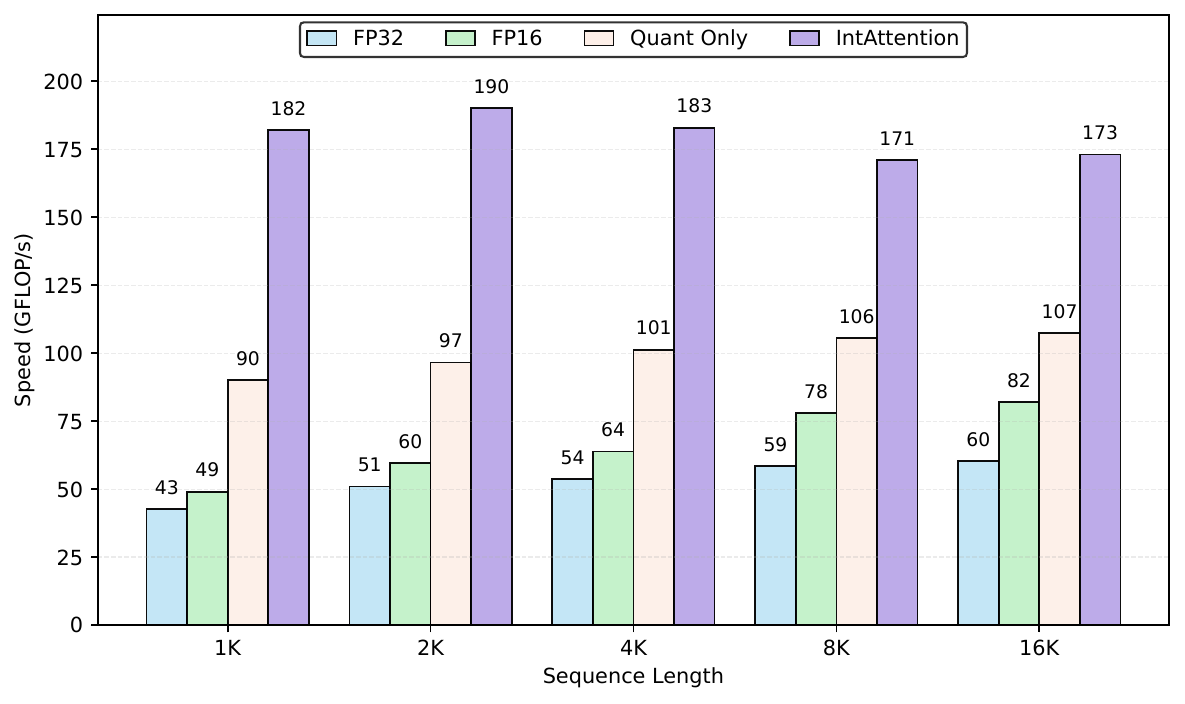}
    \vspace{-8mm}
    \caption{Speed comparison among different attention implementations on \textbf{RK3588S2} across varying sequence lengths with $d_{\text{head}}=128$.}
    \label{fig:rk3588_speed}
  \end{minipage}\hfill
  \begin{minipage}[t]{0.48\textwidth}
    \centering
    \includegraphics[width=\linewidth]{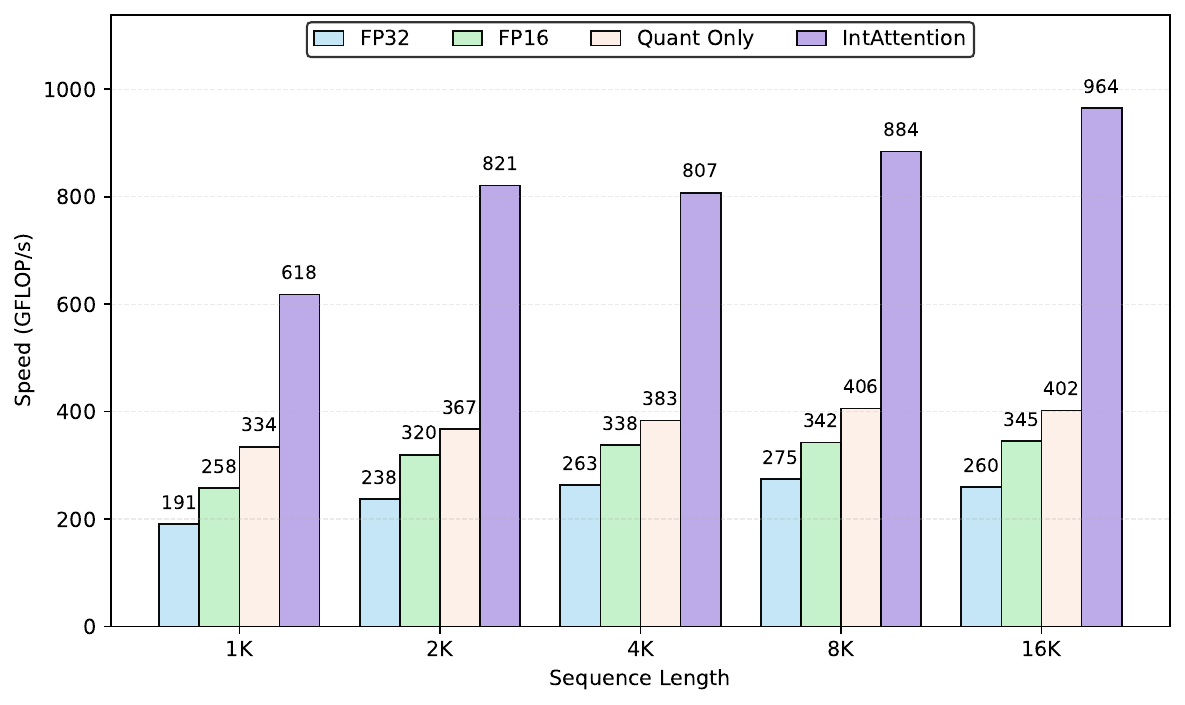}
    \vspace{-8mm}
    \caption{Speed comparison among different attention implementations on \textbf{Apple M2} across varying sequence lengths with $d_{\text{head}}=128$.}
    \label{fig:m2_speed}
  \end{minipage}
  \vspace{-1mm}
\end{figure*}

\begin{equation}
  \label{eq:lut16}
  \hat{\mathrm{LUT}}
  =
  \Big\lfloor 255 \times \mathrm{LUT} \Big\rceil,
\end{equation}
so that very small values are preserved with fine integer granularity. Given the clipped index vector $\boldsymbol{\mathrm{idx}}$, we gather a rowwise surrogate
\begin{equation}
  \label{eq:gather16}
  \hat{\mathbf{E}} = \hat{\mathrm{LUT}}\big[\boldsymbol{\mathrm{idx}}\big],
\end{equation}
accumulate its row sum with a widened integer accumulator and produce 8-bit probabilities by fixed-point scaling:
\begin{equation}
  \label{eq:uint8-prob}
  \hat{\mathbf{P}} = \Big\lfloor \frac{255 \cdot \hat{\mathbf{E}}}{\operatorname{rowSum}(\hat{\mathbf{E}})}\Big\rceil.
\end{equation}

All steps are fully integer-friendly: a single LUT gather, one 32-bit accumulation, and an elementwise scale.
By performing normalization in the integer domain, we avoid any floating-point operations in the runtime path.
As shown in \autoref{fig:exaq_vs_idxsoftmax}, compared with EXAQ, which encodes only 8 exponential values using INT3 LUT resolution under a 32-byte budget,
our \emph{IndexSoftmax} stores 32 entries within the same memory footprint,
achieving 4x higher resolution and substantially improving LUT-based approximation fidelity without requiring dynamic clipping or global statistics, which are costly on edge processors.

\subsection{Quantization Scope and Compatibility}
\label{sec:quant-scope}
Our default configuration uses per-tensor symmetric quantization for the surrounding multiplications $\mathbf{QK}^\top$ and $\mathbf{PV}$, which provides a balance between accuracy and implementation simplicity. The proposed \textbf{IndexSoftmax} is, however, compatible with finer-grained schemes such as per-channel or per-block quantization. In these cases, clipping becomes group-specific while the subsequent lookup and normalization remain unchanged.

Let the quantization be defined over groups $g=1,\dots,G$ (channels or blocks). Denote the scales by $s_Q^{(g)}$ and $s_K^{(g)}$, and define
\begin{equation}
  \alpha^{(g)} \;=\; \frac{s_Q^{(g)}\,s_K^{(g)}}{\sqrt{d}},
  \qquad
  c_{\text{int}}^{(g)} \;=\; \Big\lfloor\frac{c}{\alpha^{(g)}}\Big\rceil.
\end{equation}
Clipping in \autoref{sec:clipping} is then applied group-wise using $c_{\text{int}}^{(g)}$:
\begin{equation}
  \hat{\boldsymbol{\Delta}}^{\prime\,(g)} \;=\; \min\!\big(\hat{\boldsymbol{\Delta}}^{(g)},\, c_{\text{int}}^{(g)}\big).
\end{equation}
The index mapping and lookup table follow the same formulas with $c_{\text{int}}^{(g)}$, while the LUT itself can be shared across groups since the continuous bound $c$ and resolution $b$ are fixed:
\begin{equation}
  \boldsymbol{\mathrm{idx}}^{(g)} \;=\; \Big\lfloor\hat{\boldsymbol{\Delta}}^{\prime\,(g)}\,\frac{2^{b}-1}{c_{\text{int}}^{(g)}}\Big\rceil,\qquad
  \hat{\boldsymbol{\mathrm{E}}}^{(g)} = \hat{\mathrm{LUT}}\big[\boldsymbol{\mathrm{idx}}^{(g)}\big].
\end{equation}
Row-wise normalization in \autoref{eq:uint8-prob} proceeds identically after concatenating or summing contributions within each row. Thus, moving from per-tensor to per-channel or per-block quantization increases only the bookkeeping of scales and the computation of group-specific $c_{\text{int}}^{(g)}$, while preserving the integer-only dataflow, the LUT resolution, and the overall pipeline structure.

\begin{figure}[t]
  \centering
  \includegraphics[width=0.48\textwidth]{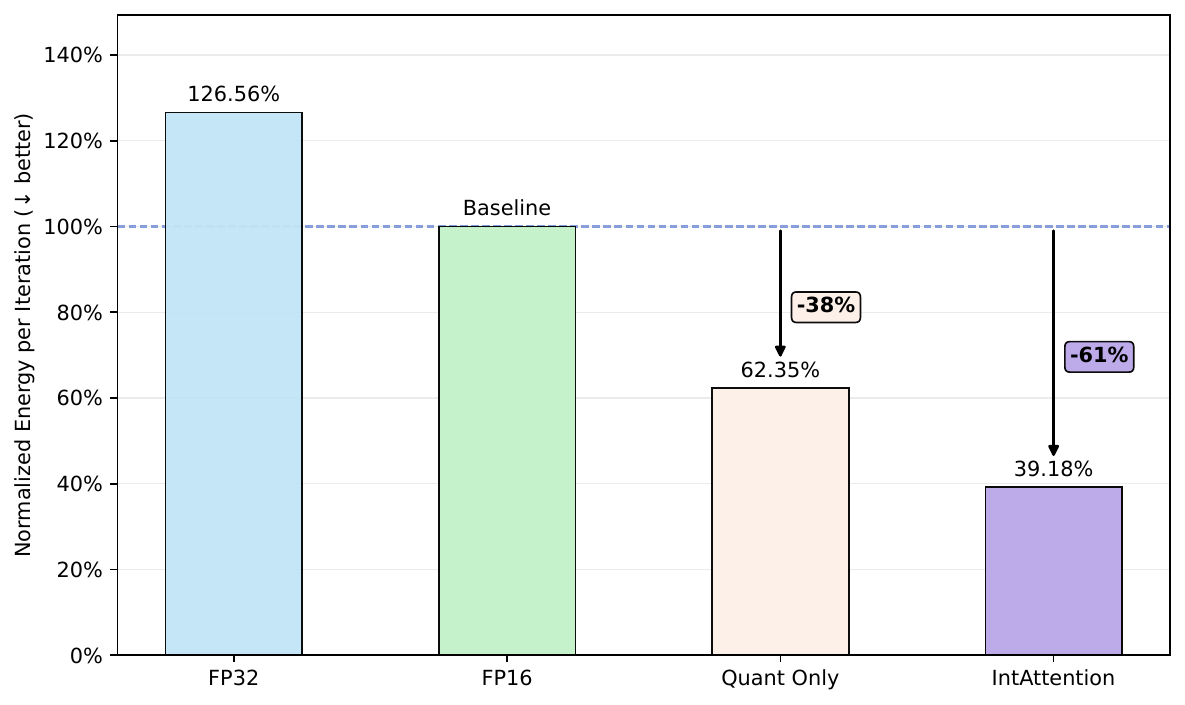}
  \vspace{-6mm}
  \caption{Normalized energy consumption per iteration across different precision settings, using FP16 as the baseline for comparison.}
  \label{fig:energy_efficiency}
\end{figure}

\begin{table*}[!t]
  \centering
  \vspace{-2mm}
  \caption{End-to-end language benchmark results for \textbf{IntAttention} compared with baselines.}
  \label{tab:intattn_acc_lang_results}
  \setlength{\tabcolsep}{4.5pt}
  \renewcommand{\arraystretch}{1.2}

  \resizebox{\textwidth}{!}{%
  \begin{tabular}{l|l|c|cccccc|c}
    \toprule
    \textbf{Model} & \textbf{Method} &
    \textbf{WikiText} $\downarrow$ &
    \textbf{HellaSwag} &
    \textbf{LAMBADA} &
    \textbf{PIQA} &
    \textbf{WinoGrande} &
    \textbf{ARC-C} &
    \textbf{ARC-E} &
    \textbf{Avg.} $\uparrow$ \\
    \midrule

    \multirow{3}{*}{\makecell{\textbf{Llama}\\\textbf{-3.2-1B}}}
    & FP16      & 12.663 & 63.65\% & 62.95\% & 74.59\% & 60.69\% & 36.18\% & 60.48\% & 59.76\% \\
    \cline{2-10}
    & Quant-Only & 13.701 & 63.39\% & 62.62\% & 74.32\% & 60.62\% & 35.84\% & \textbf{60.56\%} & 59.56\% \\
    & \cellcolor{gray!15}\textbf{IntAttention}
    & \cellcolor{gray!15}\textbf{13.070}
    & \cellcolor{gray!15}\textbf{63.50\%}
    & \cellcolor{gray!15}\textbf{63.61\%}
    & \cellcolor{gray!15}\textbf{74.92\%}
    & \cellcolor{gray!15}\textbf{61.01\%}
    & \cellcolor{gray!15}\textbf{36.43\%}
    & \cellcolor{gray!15}60.48\%
    & \cellcolor{gray!15}\textbf{59.92\%} \\
    \midrule

    \addlinespace[10pt]
    \midrule
    \multirow{3}{*}{\makecell{\textbf{OPT}\\\textbf{-1.3B}}}
    & FP16      & 16.413 & 53.73\% & 57.85\% & 72.41\% & 59.43\% & 29.61\% & 50.97\% & 54.00\% \\
    \cline{2-10}
    & Quant-Only & 18.323 & \textbf{54.11\%} & 58.00\% & \textbf{72.31\%} & 58.64\% & \textbf{29.69}\% & \textbf{50.97\%} & 53.95\% \\
    & \cellcolor{gray!15}\textbf{IntAttention}
    & \cellcolor{gray!15}\textbf{16.802}
    & \cellcolor{gray!15}53.65\%
    & \cellcolor{gray!15}\textbf{58.78\%}
    & \cellcolor{gray!15}72.25\%
    & \cellcolor{gray!15}\textbf{59.35\%}
    & \cellcolor{gray!15}29.18\%
    & \cellcolor{gray!15}50.80\%
    & \cellcolor{gray!15}\textbf{54.00\%} \\
    \midrule
    \addlinespace[10pt]
    \midrule
    \multirow{3}{*}{\makecell{\textbf{Qwen3}\\\textbf{-1.7B}}}
    & FP16     & 23.013 & 60.43\% & 50.73\% & 72.25\% & 61.09\% & 42.83\% & 69.61\% & 59.49\% \\
    \cline{2-10}
    & Quant-Only & 32.945 & \textbf{58.95\%} & 46.24\% & \textbf{68.17\%} & \textbf{58.72\%} & \textbf{37.28\%} & 61.32\% & \textbf{55.12\%} \\
    & \cellcolor{gray!15}\textbf{IntAttention}
    & \cellcolor{gray!15}\textbf{27.751}
    & \cellcolor{gray!15}58.44\%
    & \cellcolor{gray!15}\textbf{46.94\%}
    & \cellcolor{gray!15}67.30\%
    & \cellcolor{gray!15}58.56\%
    & \cellcolor{gray!15}37.03\%
    & \cellcolor{gray!15}\textbf{61.87\%}
    & \cellcolor{gray!15}55.02\% \\
    \bottomrule
  \end{tabular}}

\end{table*}

\begin{table*}[!t]
  \centering
  \caption{End-to-end vision benchmark results for \textbf{IntAttention} compared with baselines.}
  \label{tab:intattn_acc_vision_results}
  \setlength{\tabcolsep}{4.5pt}
  \renewcommand{\arraystretch}{1.2}
  \begin{tabular}{l|cc|cc|cc|cc}
    \toprule
    \multirow{2}{*}{\textbf{Method}} &
    \multicolumn{2}{c|}{\textbf{DeiT-B-224}} &
    \multicolumn{2}{c|}{\textbf{ViT-L-P16-384}} &
    \multicolumn{2}{c|}{\textbf{CaiT-L-M48-448}} &
    \multicolumn{2}{c}{\textbf{Avg.$\uparrow$}} \\
    & \textbf{Top-1} & \textbf{Top-5} & \textbf{Top-1} & \textbf{Top-5} & \textbf{Top-1} & \textbf{Top-5} & \textbf{Top-1} & \textbf{Top-5} \\
    \midrule
    FP16
    & 81.802 & 95.598 & 85.628 & 97.782 &  86.090 & 97.588 & 84.507 & 96.989 \\
    \hline
    Quant-Only
    & \textbf{81.896} & \textbf{95.708} & 83.844 & 97.150 & 85.742 & 97.530 & 83.707 & 96.796 \\
    \rowcolor{gray!15}\textbf{IntAttention}
    & 81.826 & 95.62
    & \textbf{85.224} & \textbf{97.668}
    & \textbf{86.100} & \textbf{97.640}
    & \textbf{84.383} & \textbf{96.976} \\
    \bottomrule
  \end{tabular}
\end{table*}

\begin{table*}[!t]
  \centering
  \caption{End-to-end robustness benchmark results for \textbf{IntAttention} compared with baselines. The left block reports long-context perplexity for \textbf{Llama-3.2-1B}, while the right block reports reasoning and instruction-following results for \textbf{Llama-3.2-1B-Instruct}.}
  \label{tab:robustness_attention}
  \setlength{\tabcolsep}{4.5pt}
  \renewcommand{\arraystretch}{1.2}
  \begin{tabular}{l|ccc|cccc|c}
    \toprule
    \multirow{2}{*}{\textbf{Method}}
    & \multicolumn{3}{c|}{\textbf{Llama-3.2-1B}}
    & \multicolumn{5}{c}{\textbf{Llama-3.2-1B-Instruct}} \\
    & \textbf{C4} $\downarrow$ & \textbf{OWT-10k} $\downarrow$ & \textbf{RedPajama} $\downarrow$
    & \textbf{HumanEval} & \textbf{MBPP} & \textbf{GSM8K} & \textbf{IFEval} & \textbf{Avg.} $\uparrow$ \\
    \midrule
    FP16 & 29.935 & 11.5023 & 26.756 & 32.93 & 33.00 & 33.81 & 43.44 & 35.80 \\
    \hline
    Quant-Only & 32.430 & 12.7931 & 32.189 & \textbf{32.32} & 31.40 & 34.49 & \textbf{41.40} & 34.90 \\
    \rowcolor{gray!15}
    \textbf{IntAttention} & \textbf{31.190} & \textbf{12.3178} & \textbf{28.496} & 31.10 & \textbf{34.20} & \textbf{35.03} & 39.74 & \textbf{35.02} \\
    \bottomrule
  \end{tabular}
  \vspace{-2mm}
\end{table*}

\section{Experiments}
\paragraph{Main results.}
The speed of \textbf{IntAttention} is up to \textbf{3.7$\times$ faster} than FP16 and \textbf{2.0$\times$ faster} than Quant-Only pipelines on Armv8 CPUs. Furthermore, it achieves an average \textbf{61\% energy reduction} while maintaining \textbf{strong overall fidelity}, outperforming Quant-Only and remaining close to FP16 on most evaluated language and vision settings.

\begin{figure*}[t]
  \centering
  \subfloat[Llama-3.2-1B on WikiText (PPL $\downarrow$)\label{fig:hyper_ppl}]{
    \includegraphics[width=0.48\textwidth]{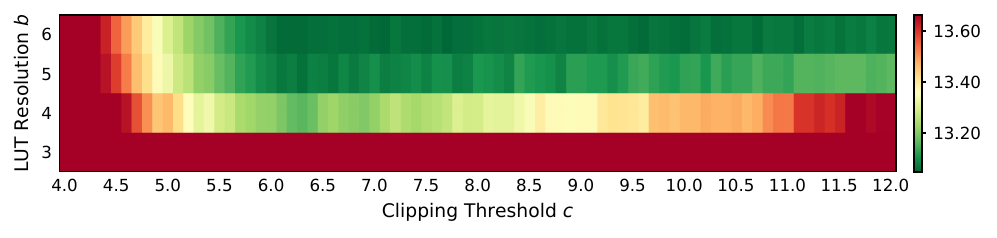}
  }\hfill
  \subfloat[DeiT-B on ImageNet-1K (Top-1 $\uparrow$)\label{fig:hyper_acc}]{
    \includegraphics[width=0.48\textwidth]{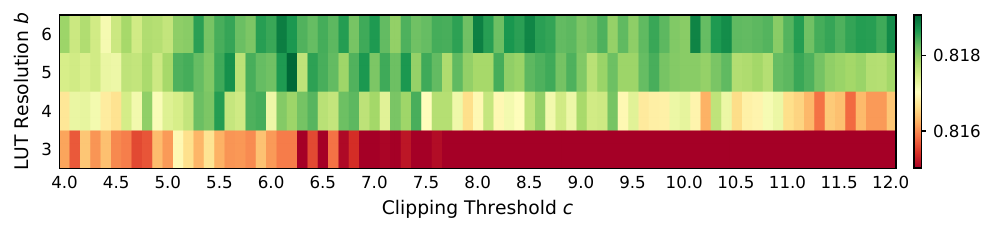}
  }
  \caption{\textbf{Hyperparameter sensitivity of IntAttention} over LUT resolution $b$ and clipping threshold $c$.
    Red indicates noticeable degradation ($>1$\,PPL or $>0.3\%$ Top-1), while green denotes high-fidelity regions.
  }
  \label{fig:hyper_joint}
\end{figure*}

\subsection{Experimental Setup}
\label{sec:setup}
\paragraph{Models.}
We evaluate \textbf{IntAttention} across both language and vision models to verify its generality.
For language, we adopt Llama-3.2-1B \cite{llama3}, OPT-1.3B \cite{zhang2022optopenpretrainedtransformer}, and Qwen3-1.7B \cite{yang2025qwen3}.
For vision, we include DeiT-B-224 \cite{deit-B}, ViT-L-P16-384 \cite{dosovitskiy2020image}, and CaiT-L-M48-448 \cite{touvron2021going}.
All models are configured under identical settings to ensure a fair comparison among different pipelines.

\paragraph{Datasets.}

For end-to-end language evaluation, we test perplexity on WikiText \cite{merity2017pointer} and report accuracy on HellaSwag \cite{zellers2019hellaswag}, LAMBADA \cite{paperno2016lambada}, PIQA \cite{Bisk2020piqa}, WinoGrande \cite{sakaguchi2021winogrande}, ARC-Challenge, and ARC-Easy \cite{allenai:arc}. To further validate robustness on more demanding tasks, we additionally evaluate Llama-3.2-1B-Instruct on GSM8K \cite{cobbe2021training}, HumanEval \cite{chen2021evaluating}, MBPP \cite{austin2021program}, and IFEval \cite{zhou2023instruction}, and evaluate Llama-3.2-1B on long-context corpora including C4 \cite{dodge2021documenting}, OpenWebText-10k \cite{Gokaslan2019OpenWeb}, and RedPajama \cite{weber2024redpajama}. Vision models are evaluated on ImageNet-1K \cite{deng2009imagenet} using the standard validation split.

\paragraph{Metrics.}
For language tasks, we report perplexity on language modeling corpora and accuracy or pass rate on downstream benchmarks, including commonsense reasoning and code generation tasks, using the open-source \emph{lm-evaluation-harness} \cite{eval-harness}. We further report the arithmetic average of accuracy-based metrics for overall comparison. For vision tasks, we report Top-1 and Top-5 accuracy on ImageNet. For efficiency evaluation, we report throughput in GFLOP/s, end-to-end attention latency in milliseconds, energy consumption per iteration, and operator-level latency breakdown. These metrics jointly characterize the trade-off between model fidelity and practical deployment efficiency.

\paragraph{Hardware and settings.}
Performance measurements are conducted on two ARMv8-based platforms, each using 8 threads.
The first is an embedded development board powered by a Rockchip RK3588S2 processor, which provides NEON and dot-product instruction support.
The second is a laptop equipped with an Apple M2 chip, supporting NEON, dot-product, and I8MM instructions.
Both systems run Arm Compute Library (ACL) version 52.4.0.
Unless otherwise specified, the sequence length $L$ takes values from ${1\mathrm{K}, 2\mathrm{K}, 4\mathrm{K}, 8\mathrm{K}, 16\mathrm{K}}$ with a typical head dimension of $d=128$.
We compare four pipelines: FP32, FP16, INT8 Quant-Only, and \emph{IntAttention}, with FP16 serving as the baseline for normalization.
\begin{table*}[!t]
\centering
\caption{
Vision benchmark results.
The upper block compares different softmax approximations,
while the lower block evaluates the complete integer attention pipeline.
}
\label{tab:combined_vision_results}

\setlength{\tabcolsep}{4.5pt}
\renewcommand{\arraystretch}{1.2}

\begin{tabular}{l|cc|cc|cc|cc}
\toprule

\multirow{2}{*}{\textbf{Method}} &
\multicolumn{2}{c|}{\textbf{DeiT-B-224}} &
\multicolumn{2}{c|}{\textbf{ViT-L-P16-384}} &
\multicolumn{2}{c|}{\textbf{CaiT-L-M48-448}} &
\multicolumn{2}{c}{\textbf{Avg.$\uparrow$}} \\

& \textbf{Top-1} & \textbf{Top-5}
& \textbf{Top-1} & \textbf{Top-5}
& \textbf{Top-1} & \textbf{Top-5}
& \textbf{Top-1} & \textbf{Top-5} \\

\midrule

FP16
& 81.802 & 95.598
& 85.628 & 97.782
& 86.090 & 97.588
& 84.507 & 96.989 \\

\cmidrule(lr){1-9}

EXAQ (INT2)
& 81.554 & 95.482
& 85.222 & 97.668
& 85.866 & 97.554
& 84.214 & 96.901 \\

EXAQ (INT3)
& 81.768 & 95.584
& 85.428 & 97.722
& 85.998 & \textbf{97.596}
& 84.398 & 96.962 \\

\rowcolor{gray!15}\textbf{IndexSoftmax}
& \textbf{81.804} & \textbf{95.590}
& \textbf{85.616} & \textbf{97.774}
& \textbf{86.114} & 97.582
& \textbf{84.511} & \textbf{96.982} \\

\cmidrule(lr){1-9}
\cmidrule(lr){1-9}

Quant-Only
& \textbf{81.896} & \textbf{95.708}
& 83.844 & 97.150
& 85.742 & 97.530
& 83.707 & 96.796 \\

\rowcolor{gray!15}\textbf{IntAttention}
& 81.826 & 95.620
& \textbf{85.224} & \textbf{97.668}
& \textbf{86.100} & \textbf{97.640}
& \textbf{84.383} & \textbf{96.976} \\

\bottomrule
\end{tabular}
\end{table*}

\subsection{Efficiency}
\label{sec:efficiency}

\begin{table*}[!t]
  \centering
  \caption{Ablation results for different softmax implementations on language benchmarks, comparing \textbf{IndexSoftmax} with EXAQ variants (INT2/INT3).}
  \label{tab:softmax_acc_lang_results}
  \setlength{\tabcolsep}{4.5pt}
  \renewcommand{\arraystretch}{1.2}

  \resizebox{\textwidth}{!}{%
  \begin{tabular}{l|l|c|cccccc|c}
    \toprule
    \textbf{Model} & \textbf{Method} &
    \textbf{WikiText} $\downarrow$ &
    \textbf{HellaSwag} &
    \textbf{LAMBADA} &
    \textbf{PIQA} &
    \textbf{WinoGrande} &
    \textbf{ARC-C} &
    \textbf{ARC-E} &
    \textbf{Avg.} $\uparrow$ \\
    \midrule

    \multirow{4}{*}{\makecell{\textbf{Llama}\\\textbf{-3.2-1B}}}
    & FP16      & 12.663 & 63.65\% & 62.95\% & 74.59\% & 60.69\% & 36.18\% & 60.48\% & 59.76\% \\
    \cline{2-10}
    & EXAQ (INT2)          & 17.753 & 57.56\% & 50.48\% & 70.73\% & 56.99\% & 33.28\% & 56.19\% & 54.21\% \\
    & EXAQ (INT3)          & 13.757 & 62.72\% & 60.72\% & 72.96\% & 58.01\% & 36.01\% & 59.55\% & 58.33\% \\
    & \cellcolor{gray!15}\textbf{IndexSoftmax}
    & \cellcolor{gray!15}\textbf{12.784}
    & \cellcolor{gray!15}\textbf{63.44\%}
    & \cellcolor{gray!15}\textbf{63.38\%}
    & \cellcolor{gray!15}\textbf{74.16\%}
    & \cellcolor{gray!15}\textbf{60.46\%}
    & \cellcolor{gray!15}\textbf{36.43\%}
    & \cellcolor{gray!15}\textbf{60.65\%}
    & \cellcolor{gray!15}\textbf{59.75\%}\\
    \midrule

    \addlinespace[10pt]
    \midrule
    \multirow{4}{*}{\makecell{\textbf{OPT}\\\textbf{-1.3B}}}
    & FP16      & 16.413 & 53.73\% & 57.85\% & 72.41\% & 59.43\% & 29.61\% & 50.97\% & 54.00\% \\
    \cline{2-10}
    & EXAQ (INT2)          & 38.782 & 52.58\% & \textbf{58.76\%} & 72.25\% & 58.33\% & 28.16\% & 50.59\% & 53.50\% \\
    & EXAQ (INT3)          & 17.248 & 53.36\% & 58.28\% & 71.93\% & \textbf{59.43\%} & \textbf{30.38\%} & 50.51\% & 53.98\% \\
    & \cellcolor{gray!15}\textbf{IndexSoftmax}
    & \cellcolor{gray!15}\textbf{16.424}
    & \cellcolor{gray!15}\textbf{53.79\%}
    & \cellcolor{gray!15}58.53\%
    & \cellcolor{gray!15}\textbf{72.20\%}
    & \cellcolor{gray!15}\textbf{59.43\%}
    & \cellcolor{gray!15}29.44\%
    & \cellcolor{gray!15}\textbf{51.01\%}
    & \cellcolor{gray!15}\textbf{54.06\%}\\
    \midrule
    \addlinespace[10pt]
    \midrule
    \multirow{4}{*}{\makecell{\textbf{Qwen3}\\\textbf{-1.7B}}}
    & FP16      & 23.013 & 60.43\% & 50.73\% & 72.25\% & 61.09\% & 42.83\% & 69.61\% & 59.49\% \\
    \cline{2-10}
    & EXAQ (INT2)          & 34.431 & 58.11\% & 45.37\% & 68.17\% & 59.35\% & 39.42\% & 63.05\% & 55.58\% \\
    & EXAQ (INT3)          & 29.006 & 59.98\% & 49.45\% & 70.78\% & \textbf{61.79\%} & 41.47\% & 67.76\% & 58.77\% \\
    & \cellcolor{gray!15}\textbf{IndexSoftmax}
    & \cellcolor{gray!15}\textbf{22.356}
    & \cellcolor{gray!15}\textbf{60.41\%}
    & \cellcolor{gray!15}\textbf{51.66\%}
    & \cellcolor{gray!15}\textbf{72.25\%}
    & \cellcolor{gray!15}61.09\%
    & \cellcolor{gray!15}\textbf{42.24\%}
    & \cellcolor{gray!15}\textbf{69.07\%}
    & \cellcolor{gray!15}\textbf{59.45\%}\\
    \bottomrule
  \end{tabular}}
\end{table*}

\begin{table*}[!t]
  \centering
  \caption{Ablation results for different softmax implementations on vision benchmarks, comparing \textbf{IndexSoftmax} with EXAQ variants (INT2/INT3).}
  \label{tab:softmax_acc_vision_results}
  \setlength{\tabcolsep}{4.5pt}
  \renewcommand{\arraystretch}{1.2}
  \begin{tabular}{l|cc|cc|cc|cc}
    \toprule
    \multirow{2}{*}{\textbf{Method}} &
    \multicolumn{2}{c|}{\textbf{DeiT-B-224}} &
    \multicolumn{2}{c|}{\textbf{ViT-L-P16-384}} &
    \multicolumn{2}{c|}{\textbf{CaiT-L-M48-448}} &
    \multicolumn{2}{c}{\textbf{Avg.$\uparrow$}} \\
    & \textbf{Top-1} & \textbf{Top-5} & \textbf{Top-1} & \textbf{Top-5} & \textbf{Top-1} & \textbf{Top-5} & \textbf{Top-1} & \textbf{Top-5} \\
    \midrule
    FP16
    & 81.802 & 95.598 & 85.628 & 97.782 &  86.090 & 97.588 & 84.507 & 96.989 \\
    \hline
    EXAQ (INT2)
    & 81.554 & 95.482 & 85.222 & 97.668 & 85.866 & 97.554 & 84.214 & 96.901 \\
    EXAQ (INT3)
    & 81.768 & 95.584 & 85.428 & 97.722 & 85.998 & \textbf{97.596} & 84.398 & 96.962 \\
    \rowcolor{gray!15}\textbf{IndexSoftmax}
    & \textbf{81.804} & \textbf{95.590}
    & \textbf{85.616} & \textbf{97.774}
    & \textbf{86.114} & 97.582 & \textbf{84.511} & \textbf{96.982} \\
    \bottomrule
  \end{tabular}
\end{table*}

\begin{table*}[!t]
  \centering
  \caption{Ablation results for different softmax implementations on robustness benchmarks, comparing \textbf{IndexSoftmax} with EXAQ variants (INT2/INT3). The left block reports long-context perplexity for \textbf{Llama-3.2-1B}, while the right block reports reasoning and instruction-following results for \textbf{Llama-3.2-1B-Instruct}.}
  \label{tab:robustness_softmax}
  \setlength{\tabcolsep}{4.5pt}
  \renewcommand{\arraystretch}{1.2}
  \begin{tabular}{l|ccc|cccc|c}
    \toprule
    \multirow{2}{*}{\textbf{Method}}
    & \multicolumn{3}{c|}{\textbf{Llama-3.2-1B}}
    & \multicolumn{5}{c}{\textbf{Llama-3.2-1B-Instruct}} \\
    & \textbf{C4} $\downarrow$ & \textbf{OWT-10k} $\downarrow$ & \textbf{RedPajama} $\downarrow$
    & \textbf{HumanEval} & \textbf{MBPP} & \textbf{GSM8K} & \textbf{IFEval} & \textbf{Avg.} $\uparrow$ \\
    \midrule
    FP16 & 29.935 & 11.5023 & 26.756 & 32.93 & 33.00 & 33.81 & 43.44 & 35.80 \\
    \hline
    EXAQ (INT2) & 39.578 & 16.0992 & 44.245 & 18.29 & 9.20 & 5.99 & 36.41 & 17.47 \\
    EXAQ (INT3) & 32.413 & 12.7847 & 30.353 & 27.44 & 27.40 & 24.26 & 39.56 & 29.67 \\
    \rowcolor{gray!15}
    \textbf{IndexSoftmax} & \textbf{30.015} & \textbf{12.0913} & \textbf{27.043} & \textbf{31.10} & \textbf{34.20} & \textbf{35.02} & \textbf{39.74} & \textbf{35.02} \\
    \bottomrule
  \end{tabular}
  \vspace{-2mm}
\end{table*}

\paragraph{Speed.}
We first measure attention throughput in GFLOP/s across sequence lengths for all four pipelines. As shown in \autoref{fig:rk3588_speed} and \autoref{fig:m2_speed}, \emph{IntAttention} consistently achieves the best throughput on both RK3588S2 and Apple M2. To provide a more intuitive system-level view, we further report end-to-end attention latency in milliseconds in \autoref{tab:latency_main}. On RK3588S2, \emph{IntAttention} achieves speedups of 2.1$\times$ to 3.7$\times$ over the FP16 baseline and 1.6$\times$ to 2$\times$ over the Quant-Only pipeline. Similar trends are also observed on Apple M2, where the speedup ranges from 2.4$\times$ to 2.8$\times$ over FP16 and from 1.9$\times$ to 2.4$\times$ over Quant-Only attention. These results show that replacing the datatype-conversion-heavy softmax path with \emph{IndexSoftmax} yields consistent end-to-end acceleration across different Arm-based CPU platforms.

\begin{table*}[!t]
  \centering
  \caption{End-to-end attention latency (ms) on RK3588S2 and Apple M2 across different sequence lengths.}
  \label{tab:latency_main}
  \setlength{\tabcolsep}{4.5pt}
  \renewcommand{\arraystretch}{1.2}
  \begin{tabular}{l|ccccc|ccccc}
    \toprule
    \multirow{2}{*}{\textbf{Method}}
    & \multicolumn{5}{c|}{\textbf{RK3588S2}}
    & \multicolumn{5}{c}{\textbf{Apple M2}} \\
    & \textbf{1K} & \textbf{2K} & \textbf{4K} & \textbf{8K} & \textbf{16K}
    & \textbf{1K} & \textbf{2K} & \textbf{4K} & \textbf{8K} & \textbf{16K} \\
    \midrule
    FP32        & 12.57 & 42.09 & 160.11 & 586.86 & 2279.96 & 2.81 & 9.04 & 32.64 & 125.13 & 529.40 \\
    FP16        & 10.96 & 36.04 & 134.64 & 440.30 & 1677.49 & 2.08 & 6.72 & 25.42 & 100.36 & 398.39 \\
    Quant-Only  & 5.96  & 22.24 & 84.85  & 325.48 & 1279.94 & 1.61 & 5.85 & 22.41 & 84.60 & 341.70 \\
    \rowcolor{gray!15}
    \textbf{IntAttention} & \textbf{2.95} & \textbf{11.29} & \textbf{46.96} & \textbf{200.94} & \textbf{794.36} & \textbf{0.87} & \textbf{2.62} & \textbf{10.64} & \textbf{38.86} & \textbf{142.50} \\
    \bottomrule
  \end{tabular}
  \vspace{-2mm}
\end{table*}

\paragraph{Energy consumption.}
We next examine energy efficiency, since edge deployment is typically power-limited.
On RK3588S2, energy per attention iteration (joules) is normalized to the FP16 baseline (\autoref{fig:energy_efficiency}). \emph{IntAttention} uses only 39.18\% of FP16 energy, a \textbf{61\%} reduction, and is \textbf{37\%} lower than the Quant-Only pipeline.
The savings come from an integer dataflow that removes dequantization, reduces memory traffic, and replaces exponentials with a compact lookup table, which translates to consistent energy benefits across all tested sequence lengths.

\subsection{Accuracy and Robustness}
We evaluate the end-to-end quality of \emph{IntAttention} under fully integer attention execution on both standard benchmarks and more challenging settings. We integrate \emph{IndexSoftmax} with quantized $\mathbf{QK}^\top$ and a UINT8 probability matrix in the $\mathbf{P}\mathbf{V}$ stage to form \emph{IntAttention}. On standard language benchmarks, as shown in \autoref{tab:intattn_acc_lang_results}, this end-to-end pipeline generally improves over the Quant-Only baseline and, for Llama-3.2-1B and OPT-1.3B, matches or slightly exceeds the FP16 baseline in average accuracy. Qwen3-1.7B remains more sensitive to attention quantization, yet \emph{IntAttention} still narrows the degradation and yields a clear perplexity improvement over the Quant-Only pipeline on WikiText. On vision benchmarks in \autoref{tab:intattn_acc_vision_results}, \emph{IntAttention} remains competitive with the Quant-Only baseline, showing slight gains on ViT-L and CaiT while staying close on DeiT-B. We further evaluate robustness on more challenging reasoning and long-context benchmarks. As shown in \autoref{tab:robustness_attention}, \emph{IntAttention} outperforms the Quant-Only baseline on all three long-context corpora and achieves stronger results on MBPP and GSM8K, yielding a slightly higher overall average. Quant-Only still retains an edge on HumanEval and IFEval, indicating that the fully integer attention pipeline preserves most of the useful attention structure even on tasks that are highly sensitive to probability distortion. Overall, these results indicate that an all-integer attention pipeline with UINT8 $\mathbf{P}$ effectively preserves probability aggregation and delivers a favorable trade-off between efficiency and accuracy across representative benchmarks and more demanding evaluation settings.

\subsection{Ablation Study}

\paragraph{Hyperparameter Sensitivity}
\emph{IntAttention} exposes only two hyperparameters: the clipping threshold $c$ and LUT resolution $b$ (see \autoref{eq:clipping_thr}, \autoref{eq:lut_re}), and both are empirically robust. A joint sweep on Llama-3.2-1B/WikiText and DeiT-B/ImageNet-1K shows a broad stability plateau for $b\!\ge\!4$ and $c\!\in[5.5,7.7]$ (\autoref{fig:hyper_joint}). Colors near the red boundary indicate insufficient LUTs or overly aggressive clipping. Intermediate tones remain acceptable, while the deep green region yields near optimal accuracy. A consistent ridge around $c{\approx}6.6$ emerges across modalities, balancing approximation fidelity with the sparsity implied by truncated exponentials. We therefore fix and recommend using $(b,c)=(5,\,6.6)$, where $b=5$ corresponds to a 32-entry UINT8 LUT ($\approx$32\,B) for all experiments with negligible memory and no measurable runtime overhead.

\paragraph{IndexSoftmax Robustness.}
We isolate the proposed softmax replacement to verify that the gains are not artifacts of model-specific tuning or benchmark choice. We replace only the softmax operator while keeping the rest of the attention pipeline unchanged, and evaluate \emph{IndexSoftmax} on standard language and vision benchmarks (\autoref{tab:softmax_acc_lang_results}, \autoref{tab:softmax_acc_vision_results}) and more challenging reasoning and long-context benchmarks (\autoref{tab:robustness_softmax}). On standard benchmarks, \emph{IndexSoftmax} preserves baseline accuracy with negligible loss and, on average, surpasses EXAQ (INT3) under the same memory budget. On reasoning and long-context tasks, it remains much closer to FP16 than EXAQ. Specifically, it yields lower perplexity on C4, OpenWebText-10k, and RedPajama, while achieving stronger results on HumanEval, MBPP, GSM8K, and IFEval. These results show that the fixed clipping and 32-entry LUT design generalizes beyond short-context classification-style evaluation, maintaining stable probability approximation under higher task complexity and longer contexts. As a result, \emph{IndexSoftmax} provides a robust and accurate integer-domain softmax replacement without per-model retuning or additional training.

\paragraph{P Matrix Quantization.}
We ablate the quantization scheme for the attention probability matrix~$\mathbf{P}$, which is constrained to the range~$[0,1]$.
As shown in \autoref{tab:p_uint8}, quantizing~$\mathbf{P}$ with a signed INT8 format causes unnecessary dynamic range waste and distorts small probability values, leading to a lower cosine similarity and higher relative L1 error.
In contrast, the unsigned UINT8 quantization fully allocates the representable range to~$[0,1]$, yielding a substantially higher similarity and lower RMSE.
This demonstrates that aligning the quantization domain with the inherent distribution of $\mathbf{P}$, a strictly positive and normalized probability matrix, is essential for maintaining attention fidelity and ensuring numerical stability in quantized attention.

\begin{table}[t]
  \centering
  \renewcommand{\arraystretch}{1.15}
  \setlength{\tabcolsep}{6pt}
  \caption{Accuracy comparison of two quantization formats for the attention probability matrix $\mathbf{P}$, evaluated against the FP16 baseline.}
  \label{tab:p_uint8}
  \begin{tabular}{l|ccc}
    \toprule
    \textbf{Format} & \textbf{CosSim} $\uparrow$ & \textbf{Relative L1} $\downarrow$ & \textbf{RMSE} $\downarrow$ \\
    \midrule
    \textbf{INT8}   & 0.996612 & 0.07739742 & 0.0023912 \\
    \textbf{UINT8}  & \textbf{0.999081} & \textbf{0.04097954} & \textbf{0.0012436} \\
    \bottomrule
  \end{tabular}
  \vspace{-3mm}
\end{table}

\paragraph{Stability Analysis.}
We further evaluate the numerical stability of the proposed integer softmax using a token-level stress test on OpenWebText with 8K context length. As shown in \autoref{tab:stability}, \emph{IndexSoftmax} exhibits no catastrophic failures: the worst-case token loss remains comparable to the FP16 baseline, the loss variance is slightly lower, and no NaN or Inf events are observed. These results indicate that the proposed clipping and lookup-table approximation remain numerically stable even under long-context stress conditions.

\begin{table}[t]
  \centering
  \caption{Stability analysis on OpenWebText with 8K context length.}
  \label{tab:stability}
  \setlength{\tabcolsep}{6pt}
  \renewcommand{\arraystretch}{1.15}
  \begin{tabular}{l|cc}
    \toprule
    \textbf{Metric} & \textbf{FP16} & \textbf{IndexSoftmax} \\
    \midrule
    Max Token Loss & 26.61 & \textbf{26.50} \\
    Loss Std. Dev. & 0.3774 & \textbf{0.3737} \\
    NaN/Inf Events & 0 & \textbf{0} \\
    \bottomrule
  \end{tabular}
  \vspace{-2mm}
\end{table}

\paragraph{Latency breakdown.}
\label{sec:breakdown}
As shown in \autoref{fig:softmax_ratio}, in a conventional INT8 attention pipeline the portion outside GEMM, namely softmax together with quantization and dequantization, becomes the dominant cost and contributes about 58\% to 65\% of total latency across sequence lengths. Replacing this stage with \emph{IndexSoftmax} and keeping the path entirely in the integer domain reduces this overhead to about 14\% to 22\% and removes dequantization. As a result, the bottleneck shifts back to the $\mathbf{QK}$ and $\mathbf{PV}$ matrix multiplications, which now account for the clear majority of time. This ablation clarifies the throughput gains: \texttt{IntAttention} removes the quantization induced hotspot and restores a compute profile in which further optimization should focus on GEMM kernels.

\subsection{Discussion}
\label{sec:discussion}

\paragraph{Accuracy perspective.}
\emph{IntAttention} maintains strong overall accuracy, and the remaining gaps arise mainly from quantizing $\mathbf{Q}$, $\mathbf{K}$, and $\mathbf{V}$ rather than softmax itself. Prior work such as \emph{SageAttention} series reduces this loss with input smoothing and finer per-block quantization scales, improving the dynamic range seen by the attention maps \cite{zhang2024sageattention, zhang2024sageattention2, zhang2025sageattention3}. Our method is orthogonal: it keeps the entire attention path in the integer domain and stabilizes probability normalization. Combining smoothing and per-block scaling with the integer dataflow of \emph{IntAttention} is expected to further improve accuracy with little additional cost.

\paragraph{Efficiency perspective.}
The latency analysis shows that removing dequantization and replacing softmax with \emph{IndexSoftmax} shifts the bottleneck back to the $\mathbf{QK}$ and $\mathbf{PV}$ matrix multiplications. This makes the optimization target clear. Future speedups should focus on stronger GEMM kernels and low-bit implementations when hardware support becomes available.

\section{Conclusion}
\label{sec:conclusion}

This paper presents \textbf{IntAttention}, a fully integer attention pipeline that serves as a training-free drop-in replacement for conventional quantized attention. By introducing \textbf{IndexSoftmax}, a LUT-based integer softmax replacement with direct integer normalization, IntAttention eliminates the costly $\mathrm{dequantize}\rightarrow\mathrm{softmax}\rightarrow\mathrm{requantize}$ path and preserves an end-to-end integer dataflow from $\mathbf{QK}^\top$ to $\mathbf{PV}$. Experiments on Armv8 CPU platforms show that IntAttention achieves up to \textbf{3.7$\times$} latency reduction and \textbf{61\%} lower energy consumption compared with FP16 baselines, while consistently outperforming conventional Quant-Only attention pipelines in efficiency. At the same time, IntAttention maintains strong overall fidelity across representative language and vision models, improves over conventional Quant-Only pipelines on more challenging long-context and reasoning settings overall. These results suggest that efficient Transformer inference on commodity edge devices requires not only integer matrix multiplications, but also an integer-native attention pipeline that removes the remaining floating-point bottleneck.

\section*{Acknowledgements}
This work was partially supported by the National Natural Science Foundation of China (Grant 62476120) and the Guangdong Basic and Applied Basic Research Foundation (2025A1515010212).


\bibliography{references}

@article{li2025questa,
  title={Questa: Expanding reasoning capacity in llms via question augmentation},
  author={Li, Jiazheng and Lin, Hongzhou and Lu, Hong and Wen, Kaiyue and Yang, Zaiwen and Gao, Jiaxuan and Wu, Yi and Zhang, Jingzhao},
  journal={arXiv preprint arXiv:2507.13266},
  year={2025}
}

@article{vaswani2017attention,
  title={Attention is all you need},
  author={Vaswani, Ashish and Shazeer, Noam and Parmar, Niki and Uszkoreit, Jakob and Jones, Llion and Gomez, Aidan N and Kaiser, {\L}ukasz and Polosukhin, Illia},
  journal={Advances in neural information processing systems},
  volume={30},
  year={2017}
}

@inproceedings{kim2021bert,
  title={I-bert: Integer-only bert quantization},
  author={Kim, Sehoon and Gholami, Amir and Yao, Zhewei and Mahoney, Michael W and Keutzer, Kurt},
  booktitle={International conference on machine learning},
  pages={5506--5518},
  year={2021},
  organization={PMLR}
}

@misc{zhang2022optopenpretrainedtransformer,
      title={OPT: Open Pre-trained Transformer Language Models}, 
      author={Susan Zhang and Stephen Roller and Naman Goyal and Mikel Artetxe and Moya Chen and Shuohui Chen and Christopher Dewan and Mona Diab and Xian Li and Xi Victoria Lin and Todor Mihaylov and Myle Ott and Sam Shleifer and Kurt Shuster and Daniel Simig and Punit Singh Koura and Anjali Sridhar and Tianlu Wang and Luke Zettlemoyer},
      year={2022},
      eprint={2205.01068},
      archivePrefix={arXiv},
      primaryClass={cs.CL},
      url={https://arxiv.org/abs/2205.01068}, 
}

@article{yang2025qwen3,
  title={Qwen3 technical report},
  author={Yang, An and Li, Anfeng and Yang, Baosong and Zhang, Beichen and Hui, Binyuan and Zheng, Bo and Yu, Bowen and Gao, Chang and Huang, Chengen and Lv, Chenxu and others},
  journal={arXiv preprint arXiv:2505.09388},
  year={2025}
}

@article{llama3,
  publtype={informal},
  author={Abhimanyu Dubey and Abhinav Jauhri and Abhinav Pandey and Abhishek Kadian and Ahmad Al-Dahle and Aiesha Letman and Akhil Mathur and Alan Schelten and Amy Yang and Angela Fan and Anirudh Goyal and Anthony Hartshorn and Aobo Yang and Archi Mitra and Archie Sravankumar and Artem Korenev and Arthur Hinsvark and Arun Rao and Aston Zhang and Aurélien Rodriguez and Austen Gregerson and Ava Spataru and Baptiste Rozière and Bethany Biron and Binh Tang and Bobbie Chern and Charlotte Caucheteux and Chaya Nayak and Chloe Bi and Chris Marra and Chris McConnell and Christian Keller and Christophe Touret and Chunyang Wu and Corinne Wong and Cristian Canton Ferrer and Cyrus Nikolaidis and Damien Allonsius and Daniel Song and Danielle Pintz and Danny Livshits and David Esiobu and Dhruv Choudhary and Dhruv Mahajan and Diego Garcia-Olano and Diego Perino and Dieuwke Hupkes and Egor Lakomkin and Ehab AlBadawy and Elina Lobanova and Emily Dinan and Eric Michael Smith and Filip Radenovic and Frank Zhang and Gabriel Synnaeve and Gabrielle Lee and Georgia Lewis Anderson and Graeme Nail and Grégoire Mialon and Guan Pang and Guillem Cucurell and Hailey Nguyen and Hannah Korevaar and Hu Xu and Hugo Touvron and Iliyan Zarov and Imanol Arrieta Ibarra and Isabel M. Kloumann and Ishan Misra and Ivan Evtimov and Jade Copet and Jaewon Lee and Jan Geffert and Jana Vranes and Jason Park and Jay Mahadeokar and Jeet Shah and Jelmer van der Linde and Jennifer Billock and Jenny Hong and Jenya Lee and Jeremy Fu and Jianfeng Chi and Jianyu Huang and Jiawen Liu and Jie Wang and Jiecao Yu and Joanna Bitton and Joe Spisak and Jongsoo Park and Joseph Rocca and Joshua Johnstun and Joshua Saxe and Junteng Jia and Kalyan Vasuden Alwala and Kartikeya Upasani and Kate Plawiak and Ke Li and Kenneth Heafield and Kevin Stone and et al.},
  title={The Llama 3 Herd of Models},
  year={2024},
  cdate={1704067200000},
  journal={CoRR},
  volume={abs/2407.21783},
  url={https://doi.org/10.48550/arXiv.2407.21783}
}

@inproceedings{deit-B,
  title={Training data-efficient image transformers \& distillation through attention},
  author={Touvron, Hugo and Cord, Matthieu and Douze, Matthijs and Massa, Francisco and Sablayrolles, Alexandre and J{\'e}gou, Herv{\'e}},
  booktitle={International conference on machine learning},
  pages={10347--10357},
  year={2021},
  organization={PMLR}
}

@article{hu2024llm,
  title={I-llm: Efficient integer-only inference for fully-quantized low-bit large language models},
  author={Hu, Xing and Cheng, Yuan and Yang, Dawei and Yuan, Zhihang and Yu, Jiangyong and Xu, Chen and Zhou, Sifan},
  journal={arXiv preprint arXiv:2405.17849},
  year={2024}
}

@inproceedings{li2023vit,
  title={I-vit: Integer-only quantization for efficient vision transformer inference},
  author={Li, Zhikai and Gu, Qingyi},
  booktitle={Proceedings of the IEEE/CVF International Conference on Computer Vision},
  pages={17065--17075},
  year={2023}
}

@inproceedings{
shkolnik2024exaq,
title={{EXAQ}: Exponent Aware Quantization For {LLM}s Acceleration},
  author={Shkolnik, Moran and Fishman, Maxim and Chmiel, Brian and Ben-Yaacov, Hilla and Banner, Ron and Levy, Kfir Yehuda},
booktitle={Workshop on Machine Learning and Compression, NeurIPS 2024},
year={2024},
}

@inproceedings{kang2024turboattention,
  title={TurboAttention: Efficient attention approximation for high throughputs llm},
  author={Kang, Hao and Bharadwaj, Srikant and Hensman, James and Krishna, Tushar and R{\"u}hle, Victor and Rajmohan, Saravan},
  booktitle={Eighth Conference on Machine Learning and Systems},
year={2025},
}

@inproceedings{liu2024consmax,
  title={Consmax: Hardware-friendly alternative softmax with learnable parameters},
  author={Liu, Shiwei and Tao, Guanchen and Zou, Yifei and Chow, Derek and Fan, Zichen and Lei, Kauna and Pan, Bangfei and Sylvester, Dennis and Kielian, Gregory and Saligane, Mehdi},
  booktitle={Proceedings of the 43rd IEEE/ACM International Conference on Computer-Aided Design},
  pages={1--9},
  year={2024}
}

@inproceedings{stevens2021softermax,
  title={Softermax: Hardware/software co-design of an efficient softmax for transformers},
  author={Stevens, Jacob R and Venkatesan, Rangharajan and Dai, Steve and Khailany, Brucek and Raghunathan, Anand},
  booktitle={2021 58th ACM/IEEE Design Automation Conference (DAC)},
  pages={469--474},
  year={2021},
  organization={IEEE}
}

@article{dao2022flashattention,
  title={Flashattention: Fast and memory-efficient exact attention with io-awareness},
  author={Dao, Tri and Fu, Dan and Ermon, Stefano and Rudra, Atri and R{\'e}, Christopher},
  journal={Advances in neural information processing systems},
  volume={35},
  pages={16344--16359},
  year={2022}
}

@inproceedings{
dao2023flashattention,
title={FlashAttention: Fast and Memory-Efficient Exact Attention with {IO}-Awareness},
author={Tri Dao and Daniel Y Fu and Stefano Ermon and Atri Rudra and Christopher Re},
booktitle={Advances in Neural Information Processing Systems},
editor={Alice H. Oh and Alekh Agarwal and Danielle Belgrave and Kyunghyun Cho},
year={2022},
}

@inproceedings{jacob2018quantization,
  title={Quantization and training of neural networks for efficient integer-arithmetic-only inference},
  author={Jacob, Benoit and Kligys, Skirmantas and Chen, Bo and Zhu, Menglong and Tang, Matthew and Howard, Andrew and Adam, Hartwig and Kalenichenko, Dmitry},
  booktitle={Proceedings of the IEEE conference on computer vision and pattern recognition},
  pages={2704--2713},
  year={2018}
}

@article{team2025gemma,
  title={Gemma 3 technical report},
  author={Team, Gemma and Kamath, Aishwarya and Ferret, Johan and Pathak, Shreya and Vieillard, Nino and Merhej, Ramona and Perrin, Sarah and Matejovicova, Tatiana and Ram{\'e}, Alexandre and Rivi{\`e}re, Morgane and others},
  journal={arXiv preprint arXiv:2503.19786},
  year={2025}
}

@inproceedings{zhang2024sageattention,
  title={SageAttention: Accurate 8-Bit Attention for Plug-and-play Inference Acceleration}, 
  author={Zhang, Jintao and Wei, Jia and Zhang, Pengle and Zhu, Jun and Chen, Jianfei},
  booktitle={International Conference on Learning Representations (ICLR)},
  year={2025}
}

@inproceedings{zhang2024sageattention2,
  title={Sageattention2: Efficient attention with thorough outlier smoothing and per-thread int4 quantization},
  author={Zhang, Jintao and Huang, Haofeng and Zhang, Pengle and Wei, Jia and Zhu, Jun and Chen, Jianfei},
  booktitle={International Conference on Machine Learning (ICML)},
  year={2025}
}

@article{micikevicius2022fp8,
  title={Fp8 formats for deep learning},
  author={Micikevicius, Paulius and Stosic, Dusan and Burgess, Neil and Cornea, Marius and Dubey, Pradeep and Grisenthwaite, Richard and Ha, Sangwon and Heinecke, Alexander and Judd, Patrick and Kamalu, John and others},
  journal={arXiv preprint arXiv:2209.05433},
  year={2022}
}

@inproceedings{kwon2023efficient,
  title={Efficient memory management for large language model serving with pagedattention},
  author={Kwon, Woosuk and Li, Zhuohan and Zhuang, Siyuan and Sheng, Ying and Zheng, Lianmin and Yu, Cody Hao and Gonzalez, Joseph and Zhang, Hao and Stoica, Ion},
  booktitle={Proceedings of the 29th symposium on operating systems principles},
  pages={611--626},
  year={2023}
}

@article{shah2024flashattention,
  title={Flashattention-3: Fast and accurate attention with asynchrony and low-precision},
  author={Shah, Jay and Bikshandi, Ganesh and Zhang, Ying and Thakkar, Vijay and Ramani, Pradeep and Dao, Tri},
  journal={Advances in Neural Information Processing Systems},
  volume={37},
  pages={68658--68685},
  year={2024}
}

@inproceedings{radford2021learning,
  title={Learning transferable visual models from natural language supervision},
  author={Radford, Alec and Kim, Jong Wook and Hallacy, Chris and Ramesh, Aditya and Goh, Gabriel and Agarwal, Sandhini and Sastry, Girish and Askell, Amanda and Mishkin, Pamela and Clark, Jack and others},
  booktitle={International conference on machine learning},
  pages={8748--8763},
  year={2021},
  organization={PmLR}
}

@inproceedings{
dosovitskiy2020image,
title={An Image is Worth 16x16 Words: Transformers for Image Recognition at Scale},
author={Alexey Dosovitskiy and Lucas Beyer and Alexander Kolesnikov and Dirk Weissenborn and Xiaohua Zhai and Thomas Unterthiner and Mostafa Dehghani and Matthias Minderer and Georg Heigold and Sylvain Gelly and Jakob Uszkoreit and Neil Houlsby},
booktitle={International Conference on Learning Representations},
year={2021},
}

@article{dettmers2022gpt3,
  title={Gpt3. int8 (): 8-bit matrix multiplication for transformers at scale},
  author={Dettmers, Tim and Lewis, Mike and Belkada, Younes and Zettlemoyer, Luke},
  journal={Advances in neural information processing systems},
  volume={35},
  pages={30318--30332},
  year={2022}
}

@inproceedings{
zhang2025sageattention3,
title={SageAttention3: Microscaling {FP}4 Attention for Inference and An Exploration of 8-Bit Training},
  author={Zhang, Jintao and Wei, Jia and Zhang, Pengle and Xu, Xiaoming and Huang, Haofeng and Wang, Haoxu and Jiang, Kai and Zhu, Jun and Chen, Jianfei},
booktitle={The Thirty-ninth Annual Conference on Neural Information Processing Systems},
year={2025},
url={https://openreview.net/forum?id=JbJVWljk7r}
}

@inproceedings{Bisk2020piqa,
  author = {Yonatan Bisk and Rowan Zellers and
            Ronan Le Bras and Jianfeng Gao
            and Yejin Choi},
  title = {PIQA: Reasoning about Physical Commonsense in
           Natural Language},
  booktitle = {Thirty-Fourth AAAI Conference on
               Artificial Intelligence},
  year = {2020},
}

@inproceedings{zellers2019hellaswag,
    title={HellaSwag: Can a Machine Really Finish Your Sentence?},
    author={Zellers, Rowan and Holtzman, Ari and Bisk, Yonatan and Farhadi, Ali and Choi, Yejin},
    booktitle ={Proceedings of the 57th Annual Meeting of the Association for Computational Linguistics},
    year={2019}
}

@article{allenai:arc,
      author    = {Peter Clark  and Isaac Cowhey and Oren Etzioni and Tushar Khot and
                    Ashish Sabharwal and Carissa Schoenick and Oyvind Tafjord},
      title     = {Think you have Solved Question Answering? Try ARC, the AI2 Reasoning Challenge},
      journal   = {arXiv:1803.05457v1},
      year      = {2018},
}

@inproceedings{paperno2016lambada,
  title={The LAMBADA dataset: Word prediction requiring a broad discourse context},
  author={Paperno, D and Kruszewski, G and Lazaridou, A and Pham, QN and Bernardi, Raffaella and Pezzelle, S and Baroni, M and Boleda, G and Fern{\'a}ndez, R},
  booktitle={54th Annual Meeting of the Association for Computational Linguistics, ACL 2016-Long Papers},
  volume={3},
  pages={1525--1534},
  year={2016},
  organization={Association for Computational Linguistics (ACL)}
}

@misc{eval-harness,
  author       = {Gao, Leo and Tow, Jonathan and Abbasi, Baber and Biderman, Stella and Black, Sid and DiPofi, Anthony and Foster, Charles and Golding, Laurence and Hsu, Jeffrey and Le Noac'h, Alain and Li, Haonan and McDonell, Kyle and Muennighoff, Niklas and Ociepa, Chris and Phang, Jason and Reynolds, Laria and Schoelkopf, Hailey and Skowron, Aviya and Sutawika, Lintang and Tang, Eric and Thite, Anish and Wang, Ben and Wang, Kevin and Zou, Andy},
  title        = {The Language Model Evaluation Harness},
  month        = 07,
  year         = 2024,
  publisher    = {Zenodo},
  version      = {v0.4.3},
  doi          = {10.5281/zenodo.12608602},
  url          = {https://zenodo.org/records/12608602}
}

@inproceedings{deng2009imagenet,
  title={Imagenet: A large-scale hierarchical image database},
  author={Deng, Jia and Dong, Wei and Socher, Richard and Li, Li-Jia and Li, Kai and Fei-Fei, Li},
  booktitle={2009 IEEE conference on computer vision and pattern recognition},
  pages={248--255},
  year={2009},
  organization={Ieee}
}

@inproceedings{merity2017pointer,
  title={Pointer Sentinel Mixture Models},
  author={Merity, Stephen and Xiong, Caiming and Bradbury, James and Socher, Richard},
  booktitle={International Conference on Learning Representations},
  year={2017}
}

@article{sakaguchi2021winogrande,
  title={Winogrande: An adversarial winograd schema challenge at scale},
  author={Sakaguchi, Keisuke and Bras, Ronan Le and Bhagavatula, Chandra and Choi, Yejin},
  journal={Communications of the ACM},
  volume={64},
  number={9},
  pages={99--106},
  year={2021},
  publisher={ACM New York, NY, USA}
}

@inproceedings{touvron2021going,
  title={Going deeper with image transformers},
  author={Touvron, Hugo and Cord, Matthieu and Sablayrolles, Alexandre and Synnaeve, Gabriel and J{\'e}gou, Herv{\'e}},
  booktitle={Proceedings of the IEEE/CVF international conference on computer vision},
  pages={32--42},
  year={2021}
}

@article{cobbe2021training,
  title={Training verifiers to solve math word problems},
  author={Cobbe, Karl and Kosaraju, Vineet and Bavarian, Mohammad and Chen, Mark and Jun, Heewoo and Kaiser, Lukasz and Plappert, Matthias and Tworek, Jerry and Hilton, Jacob and Nakano, Reiichiro and others},
  journal={arXiv preprint arXiv:2110.14168},
  year={2021}
}

@article{chen2021evaluating,
  title={Evaluating large language models trained on code},
  author={Chen, Mark and Tworek, Jerry and Jun, Heewoo and Yuan, Qiming and Pinto, Henrique Ponde De Oliveira and Kaplan, Jared and Edwards, Harri and Burda, Yuri and Joseph, Nicholas and Brockman, Greg and others},
  journal={arXiv preprint arXiv:2107.03374},
  year={2021}
}

@article{austin2021program,
  title={Program synthesis with large language models},
  author={Austin, Jacob and Odena, Augustus and Nye, Maxwell and Bosma, Maarten and Michalewski, Henryk and Dohan, David and Jiang, Ellen and Cai, Carrie and Terry, Michael and Le, Quoc and others},
  journal={arXiv preprint arXiv:2108.07732},
  year={2021}
}

@article{zhou2023instruction,
  title={Instruction-following evaluation for large language models},
  author={Zhou, Jeffrey and Lu, Tianjian and Mishra, Swaroop and Brahma, Siddhartha and Basu, Sujoy and Luan, Yi and Zhou, Denny and Hou, Le},
  journal={arXiv preprint arXiv:2311.07911},
  year={2023}
}

@inproceedings{dodge2021documenting,
  title={Documenting large webtext corpora: A case study on the colossal clean crawled corpus},
  author={Dodge, Jesse and Sap, Maarten and Marasovi{\'c}, Ana and Agnew, William and Ilharco, Gabriel and Groeneveld, Dirk and Mitchell, Margaret and Gardner, Matt},
  booktitle={Proceedings of the 2021 conference on empirical methods in natural language processing},
  pages={1286--1305},
  year={2021}
}

@article{weber2024redpajama,
  title={Redpajama: an open dataset for training large language models},
  author={Weber, Maurice and Fu, Daniel Y and Anthony, Quentin and Oren, Yonatan and Adams, Shane and Alexandrov, Anton and Lyu, Xiaozhong and Nguyen, Huu and Yao, Xiaozhe and Adams, Virginia and others},
  journal={Advances in neural information processing systems},
  volume={37},
  pages={116462--116492},
  year={2024}
}

@misc{Gokaslan2019OpenWeb,
    title={OpenWebText Corpus},
    author={Gokaslan, Aaron and Cohen, Vanya and Pavlick, Ellie and Tellex, Stefanie},
    howpublished={\url{http://Skylion007.github.io/OpenWebTextCorpus}},
    year={2019}
}
\bibliographystyle{mlsys2026}


\end{document}